%% file: iclr2027_conference.tex
\documentclass{article} 
\usepackage{iclr2027_conference,times}

\input{math_commands.tex}

\usepackage{hyperref}
\usepackage{url}

\usepackage{wrapfig}
\usepackage{graphicx}
\usepackage{multirow}
\usepackage{booktabs}
\usepackage[table]{xcolor}
\usepackage{tabularx}
\usepackage{array}
\usepackage{amssymb}
\usepackage{caption}

\newcommand{\graymidrule}{%
  \arrayrulecolor{black!20}%
  \specialrule{0.4pt}{0.4pt}{1.4pt}%
  \arrayrulecolor{black}%
}

\definecolor{perfblue}{RGB}{78,149,217}

\title{Persistence Forcing: Exploiting Feature Specialization in Pixel-Space Diffusion}

\author{
Chong Wang$^1$, Zixuan Fu$^1$, Shiqi Huang$^1$, Siyuan Yang$^2$,
Hao Cheng$^3$, Bihan Wen$^{1}$%
\thanks{Corresponding author: Bihan Wen (\texttt{bihan.wen@ntu.edu.sg}).} \\
$^1$Nanyang Technological University \\
$^2$KTH Royal Institute of Technology \qquad
$^3$Hebei University of Technology\\
\texttt{wang1711@e.ntu.edu.sg}
}

\iclrfinalcopy 
\begin{document}

\maketitle

\fancyhead[L]{Preprint}

\begin{center}
\vspace{-0.4in}
\href{https://chongwang1024.github.io/PerF}{\textcolor{perfblue}{\textbf{https://chongwang1024.github.io/PerF}}}
\end{center}

\begin{abstract}

Pixel-space diffusion Transformers (DiTs) directly operate on high-dimensional visual data, yet their hidden representations typically undergo uniform refinement across depth.
%
Natural images, however, are inherently organized at different levels of granularity.
Global structure can often be represented compactly, whereas local textures and fine details require richer representations.
%
%
Motivated by this, we introduce heterogeneous refinement in pixel-space DiTs, assigning different feature groups distinct refinement budgets across depth.
Consequently, an ordered feature specialization emerges: sparsely refined features predominantly encode global visual structure, whereas more frequently refined features increasingly specialize toward localized, high-frequency details.
We refer to these two groups as persistent and active features, respectively.
Building on this emergent specialization, we introduce \emph{Persistence Forcing (PerF)}, which explicitly exploits this persistent--active feature organization for pixel-space image generation.
This enables persistent features to continuously condition actively refined features, allowing stable global information to guide the ongoing refinement of finer visual details.
During generative sampling, this interaction further induces a meaningful guidance direction that promotes coherent global structure and naturally complements classifier-free guidance.
%
%
On ImageNet $256\times256$, PerF-L achieves FID of $1.91$, approaching $1.86$ of JiT-H with only half the parameters, while PerF-H further achieves FID of $1.63$ and $1.76$ on ImageNet $256\times256$ and $512\times512$, respectively.
\end{abstract}

\begin{figure}[ht]
    \centering
    \vspace{-15pt}
    \includegraphics[width=0.82\linewidth]{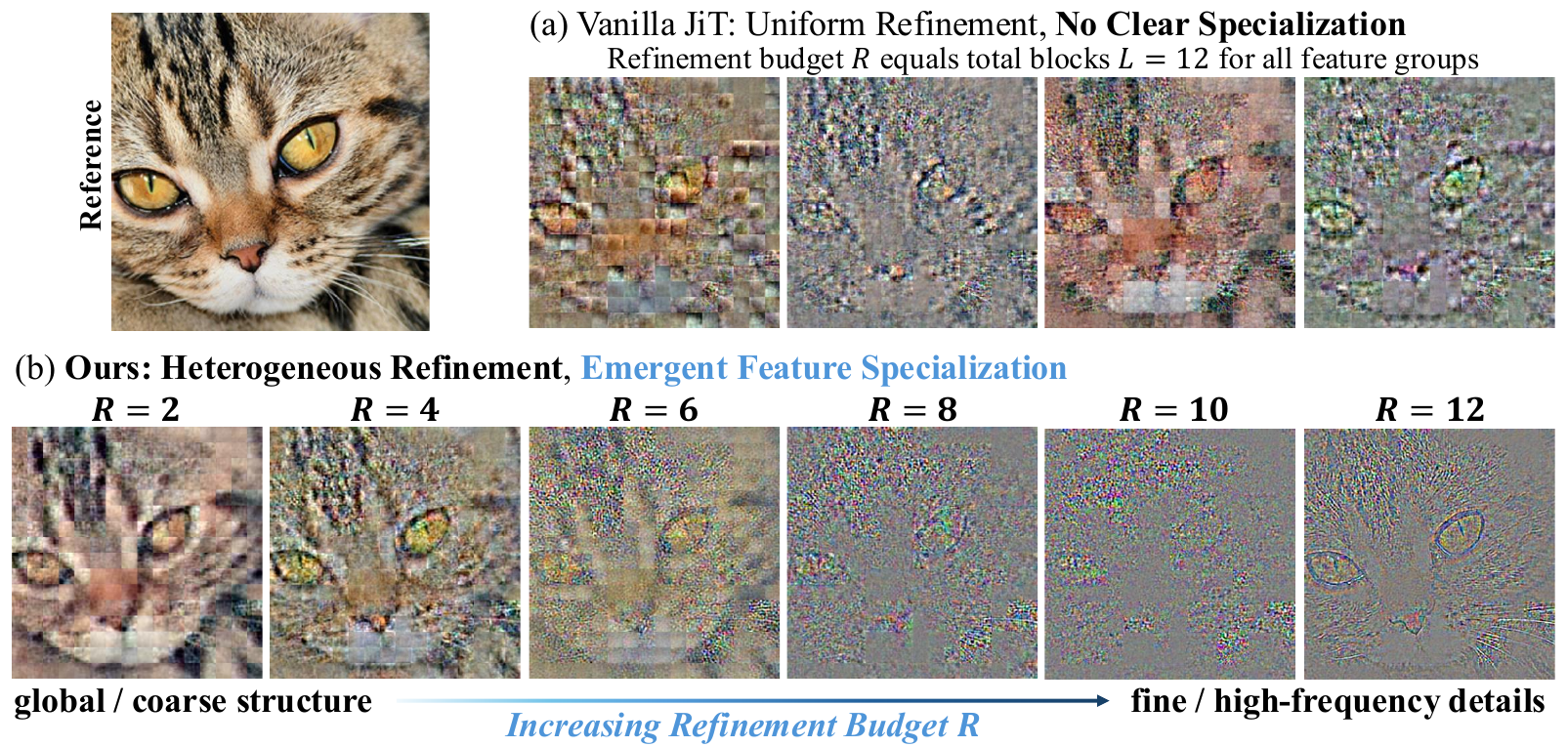}
    \vspace{-7pt}
    \caption{
    \textbf{Heterogeneous refinement induces ordered feature specialization.}
    We visualize the contributions of different feature groups to the final RGB prediction. Under uniform refinement, feature groups show no systematic ordering. Under heterogeneous refinement, increasing refinement budget $R$ organizes their contributions from global structure toward fine, high-frequency details.
    }
    \label{fig:feature_specialization}
    \vspace{-15pt}
\end{figure}

\section{Introduction}\vspace{-10pt}
Learning effective internal representations is a key bottleneck in training generative Diffusion Transformers (DiTs)~\citep{peebles2023scalable,yu2025representation,wang2026ddt}.
In particular, recent pixel-space diffusion models~\citep{li2026back,chen2026dip,yu2026pixeldit} directly operate on high-dimensional visual data without relying on a pretrained latent representation~\citep{rombach2022high,zheng2026diffusion} and therefore rely heavily on the model itself to organize useful representations throughout denoising.
Despite rapid progress in architectures~\citep{wang2026ddt}, training objectives~\citep{ma2026deco}, and sampling strategies~\citep{zhou2026guiding}, DiTs largely retain a simple computational pattern: the full hidden representation is repeatedly transformed throughout network depth.
As a result, hidden features typically undergo a uniform refinement schedule.
%
Natural images, however, are inherently organized at different levels of granularity.
Global structure can often be represented compactly, whereas local textures and fine details require richer representations~\citep{burt1983laplacian, mallat1989theory, schwartz2001natural}.
This distinction is particularly relevant to pixel-space diffusion, where a single model must jointly organize global semantic structure and resolve fine-grained, high-frequency visual details directly in the raw pixel space~\citep{chen2026dip,ma2026deco,yu2026pixeldit}.
This motivates us to reconsider whether all hidden features should receive the same amount of refinement in pixel-space DiTs.

We therefore introduce heterogeneous refinement, assigning different feature groups distinct refinement budgets across network depth.
Under this design, an ordered pattern of feature specialization emerges.
Sparsely refined features predominantly encode coherent global visual
structure, whereas more frequently refined features increasingly specialize toward localized, high-frequency details.
We refer to these two groups as \emph{persistent} and \emph{active} features, respectively.
Notably, persistent features already exhibit coherent global organization after limited refinement and remain stable while active features continue to evolve toward progressively finer details.
Heterogeneous refinement therefore gives rise to a natural persistent--active organization of the hidden representation, as illustrated in Figure~\ref{fig:feature_specialization}.
This emergent feature specialization further opens the possibility of using the persistent features as stable context for features that remain actively refined.
%


%

To this end, we propose \emph{Persistence Forcing (PerF)}, which explicitly exploits the persistent--active feature specialization induced by heterogeneous refinement.
During denoising, persistent features continuously condition actively refined representations, allowing stable global information to guide the ongoing refinement of progressively finer visual details.
This persistent-to-active interaction is related to the broader coarse-to-fine principle explored in recent diffusion architectures, where global representations condition high-frequency or pixel-level refinement~\citep{wang2026ddt,ma2026deco,yu2026pixeldit}.
Unlike these explicitly designed hierarchies, however, PerF builds on the feature specialization that emerges naturally from heterogeneous refinement.
We further leverage this interaction at sampling time through Persistence Guidance (PG).
By contrasting predictions with and without persistent conditioning, PG constructs a guidance direction that strengthens the influence of the internally established structural context.
This guidance complements the externally specified semantic signal provided by classifier-free guidance (CFG)~\citep{ho2022classifier}.
On ImageNet $256\times256$ generation, PerF consistently improves the corresponding JiT baselines across multiple model scales, reducing FID from $3.66$ to $2.81$ for B/16, from $2.36$ to $1.91$ for L/16, and from $1.86$ to $1.63$ for H/16.
The improvement further extends to ImageNet $512\times512$ generation, where PerF-H/32 reduces FID from $1.94$ to $1.76$ and achieves an Inception Score of $335.3$.

We highlight the main contribution of this paper below:
\begin{itemize}
    \item We introduce heterogeneous refinement in pixel-space diffusion Transformers and reveal an ordered form of feature specialization, where different refinement budgets naturally yield persistent and actively refined representations with complementary coarse-to-fine roles.
    \item We propose \emph{Persistence Forcing}, which leverages persistent representations as structural context for active refinement and amplifies their influence through sampling-time guidance.
    \item PerF consistently improves JiT across model scales and resolutions, reducing FID from $3.66$ to $2.81$ for JiT-B on ImageNet $256\times256$, while PerF-H achieves FIDs of $1.63$ and $1.76$ at $256\times256$ and $512\times512$ resolutions, respectively.
\end{itemize}

\section{Heterogeneous Refinement Induces Feature Specialization}
\subsection{Heterogeneous Refinement via Variable Width}
%
%

\begin{wrapfigure}{r}{0.48\textwidth}
    \centering
    \includegraphics[width=1.0\linewidth]{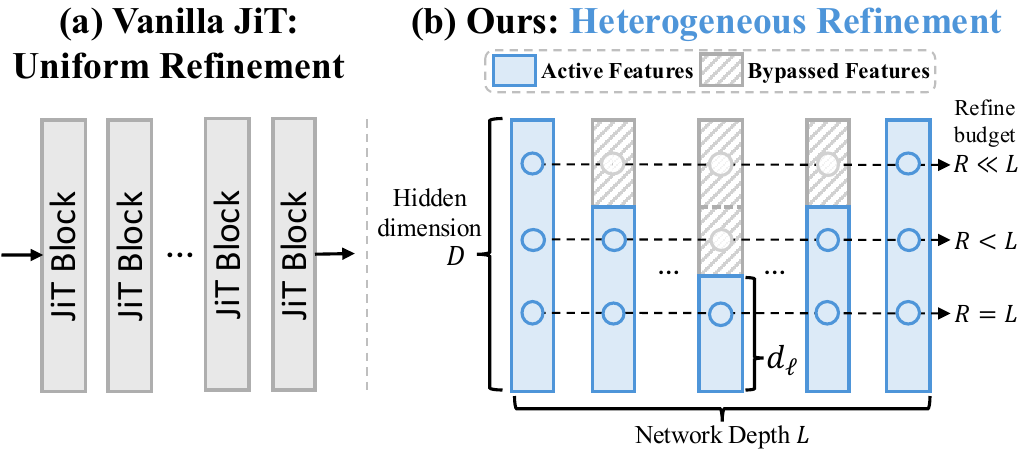}
    \vspace{-20pt}
    \caption{
\textbf{From uniform to heterogeneous refinement.}
(a) In vanilla JiT~\citep{li2026back}, every hidden dimension is updated at
every block, yielding the same refinement budget $R_i=L$.
(b) Our design updates only the first $d_\ell$ dimensions at layer $\ell$
while bypassing the remaining.
Varying $d_\ell$ across depth assigns different refinement budgets to
different feature dimensions.
    }
    \label{fig:variable-width}
    \vspace{-25pt}
\end{wrapfigure}

Existing pixel-space diffusion Transformers~\citep{li2026back,baade2026latent} largely inherit the standard DiT design~\citep{peebles2023scalable}, in which the entire hidden representation is propagated by every Transformer block.
Consequently, all hidden feature dimensions undergo the same number of refinement steps throughout network depth.
%
%
We question whether such uniform refinement is necessary for pixel-space generation, where a single end-to-end network must organize representations ranging from global visual structure to fine-grained image details.
To investigate this question, we introduce heterogeneous refinement, where different feature groups receive distinct refinement budgets across network depth.
We instantiate this study on JiT~\citep{li2026back}, a simple uniform-width pixel-space DiT that provides a clean setting for isolating the effect of heterogeneous refinement.
%
As illustrated in Figure~\ref{fig:variable-width}, we maintain a $D$-dimensional hidden representation and allow the $\ell$-th Transformer block to actively transform only the first $d_\ell \leq D$ feature dimensions.
The remaining dimensions bypass the block unchanged and can re-enter computation when the width expands in later layers.
We quantify how frequently each feature, indexed by $i$, is transformed by its refinement budget
\begin{equation}
    R_i
    =
    \sum_{\ell=1}^{L}
    \mathbf{1}[i \leq d_\ell],
\end{equation}
indicating that only first $d_\ell$ feature channels are active at layer $\ell$.
Under uniform refinement, as in vanilla JiT~\citep{li2026back}, $R_i=L$ for all feature dimensions.
By varying $d_\ell$ across depth, our construction instead creates a range of refinement budgets and thus heterogeneous refinement histories across the hidden representation.
This layer-wise width variation has recently been explored in decoder-only language models as a means of nonuniform capacity allocation across network depth ~\citep{wu2026variable}.
Here, we use this architectural degree of freedom to induce heterogeneous refinement histories across feature dimensions in pixel-space diffusion.

In this work, we utilize a contraction--expansion schedule for $d_\ell$, gradually narrowing and then expanding the active width across depth to produce a broad range of refinement budgets while allowing bypassed features to re-enter later computation.
%
%
To isolate the effect of heterogeneous refinement from changes in backbone capacity, we approximately match the attention and MLP parameter budget of the corresponding vanilla JiT by preserving $\sum_{\ell=1}^{L} d_\ell^2 = L d^2$, where $d$ denotes the hidden dimension of the corresponding JiT baseline.
%

\subsection{Emergent Feature Specialization}\label{sec_specilization}

Heterogeneous refinement assigns different feature groups distinct refinement budgets, without prescribing their representational roles.
Despite this minimal intervention, an ordered feature specialization emerges.

To make this specialization visible, we decompose the final RGB prediction into the additive contributions of different feature groups through the model output projection.
Figure~\ref{fig:feature_specialization} visualizes these contributions for groups with different refinement budgets.
Under heterogeneous refinement, they exhibit an ordered coarse-to-fine progression.
Groups receiving only limited refinement already preserve coherent object appearance and global image structure, while increasing refinement generally shifts their contributions toward more localized and higher-frequency content, including textures, boundaries, and residual details.
In contrast, feature partitions under uniform refinement all receive the same refinement budget and show no comparable ordering.
Therefore, heterogeneous refinement does more than expose different feature groups to different amounts of computation.
Moreover, it organizes their representational roles according to their refinement histories.

This organization reveals a natural feature specialization.
We refer to sparsely refined features that preserve predominantly global visual structure as \emph{persistent features}, and to features that remain repeatedly transformed throughout network depth as \emph{actively refined features}.
Notably, these roles are not explicitly assigned. The architecture determines only the refinement history of each feature group, while its representational role emerges through denoising.

\subsection{Heterogeneous Refinement Alone Is Not Enough}\label{sec_not_enough}
\begin{wraptable}{r}{0.45\textwidth}
\centering
\vspace{-10pt}
\scriptsize
\setlength{\tabcolsep}{5pt}
\begin{tabular}{
cc
cc
cc
}
\toprule
\multirow{2}{*}{Model} & \multirow{2}{*}{Params.} &
\multicolumn{2}{c}{100 Epoch} &
\multicolumn{2}{c}{200 Epoch} \\
\cmidrule(lr){3-4}
\cmidrule(lr){5-6}
& &
FID $\downarrow$ &
IS $\uparrow$ &
FID $\downarrow$ &
IS $\uparrow$ \\
\midrule

Uni.
& 131M
& \textbf{42.1}
& \textbf{37.6}
& \textbf{33.1}
& \textbf{49.8} \\
Hetero.
& 131M
& 42.7
& 36.2
& 33.8
& 47.1 \\
\bottomrule
\end{tabular}
\vspace{-5pt}
\caption{\textbf{Heterogeneous refinement alone does not readily achieve better generation.} 
}
\label{tab_vw_only}
\vspace{-12pt}
\end{wraptable}

However, the feature specialization induced by heterogeneous refinement above does not directly translate into better generation performance.
Despite exhibiting a clear coarse-to-fine feature specialization, a parameter-matched model with heterogeneous refinement does not outperform the corresponding vanilla JiT and can even produce worse FID, as shown in Table~\ref{tab_vw_only}.
This gap suggests that heterogeneous refinement provides an effective mechanism for revealing distinct representational roles, but is not sufficient by itself to improve generation quality.

%
The remaining opportunity lies in how these roles interact.
In the basic heterogeneous-refinement construction, persistent features are carried forward while actively refined features continue through the Transformer blocks, but the preserved information is not explicitly used to support the ongoing computation.
This motivates us to move beyond inducing heterogeneous refinement and explicitly exploit the persistent representations it creates.


\section{Persistence Forcing}

\subsection{Persistent-to-Active Conditioning}\label{sec:persistent_conditioning}


Section~\ref{sec_specilization} reveals a natural feature specialization under heterogeneous refinement.
Motivated by their specialization, we explicitly construct the interaction between persistent and actively refined features by using the persistent representation as a structural context for active refinement.
Rather than repeatedly transforming information that has already established coherent global organization, we preserve it and allow it to continuously inform the features that remain under active computation.
This turns persistence from passive feature preservation into an explicit interaction between the two representational roles.

At the $\ell$-th Transformer block, we denote the actively transformed features by $\mathbf{h}^{a}_{\ell}$ and the persistent features bypassing the block by $\mathbf{h}^{p}_{\ell}$.
While $\mathbf{h}^{a}_{\ell}$ is updated by the Transformer block, $\mathbf{h}^{p}_{\ell}$ is withheld from transformation at this block and instead serves as an additional modulation condition for the active features.
Specifically, we augment the standard timestep and class modulation with a token-wise modulation~\citep{wang2026ddt} derived from the persistent representation,
\begin{equation}
    \mathbf{m}_{\ell}
    =
    \operatorname{AdaLN}_{\ell}(t,y)
    +
    \mathcal{P}_{\ell}
    \left(
        \operatorname{Norm}(\mathbf{h}^{p}_{\ell})
    \right).
\end{equation}
The first term denotes the standard timestep and class modulation,
which is shared across image tokens, while the second provides
token-wise modulation derived from the persistent representation.
%
%
Specifically, the resulting persistent modulation remains spatially aligned with the active representation, allowing the established structure of the current sample to directly inform its subsequent refinement.

The predominantly coarse and globally organized information carried by persistent features also suggests that this conditioning can be represented compactly.
We therefore implement $\mathcal{P}_{\ell}$ with a low-rank bottleneck of dimension $r$, providing a lightweight structural condition for active refinement.
Persistent information can thus influence every active refinement step without requiring the persistent representation itself to be repeatedly transformed.
This interaction follows a coarse-to-fine principle related to recent decoupled diffusion architectures~\citep{wang2026ddt,ma2026deco,yu2026pixeldit}, while the two roles here emerge within a single backbone from heterogeneous refinement rather than being explicitly assigned.

Importantly, making the persistent influence explicit also turns it into a controllable internal conditioning variable.
In the next section, we exploit this property to construct Persistence Guidance.


\begin{figure}[t]
    \centering
    \includegraphics[width=0.8\linewidth]{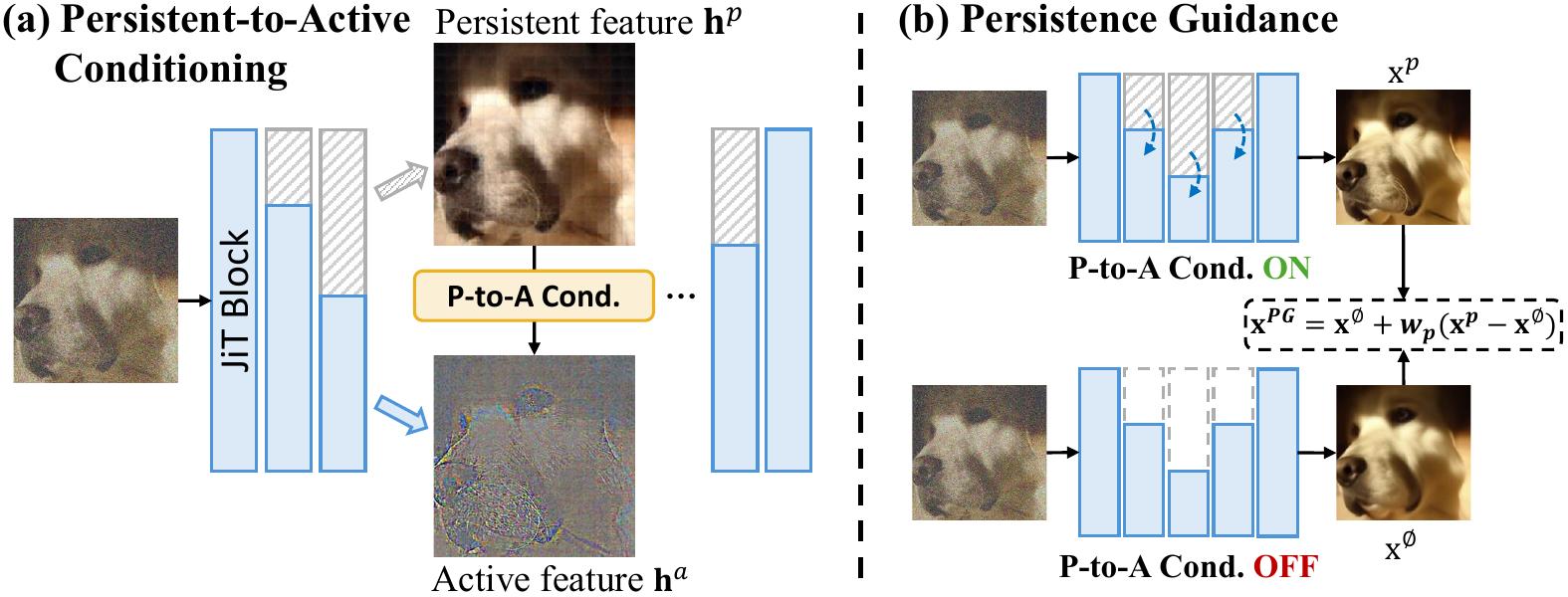}
    \captionof{figure}{
        \textbf{Overview of Persistence Forcing.}
(a) Persistent-to-active conditioning explicitly uses preserved persistent features as structural context for actively refined features. (b) The resulting controllable interaction is further exploited to construct Persistence Guidance.
    }
    \vspace{-10pt}
    \label{fig:framework}
\end{figure}

\subsection{Persistence Guidance}
%
Guidance can be interpreted as extrapolating along a meaningful contrast between two denoising predictions.
Classifier-free guidance (CFG)~\citep{ho2022classifier} forms this contrast by removing an external semantic condition, while recent self-guidance approaches construct it from intermediate predictions of the same model~\citep{zhou2026guiding}.
These methods highlight that the effectiveness of guidance depends on what information the prediction contrast isolates.

Persistent-to-active conditioning not only provides stable context for active refinement, but also exposes how persistent representations influence the denoising prediction.
This conditioning provides a different and directly interpretable contrast which can be further utilized as meaningful guidance during the generative sampling.
We therefore stochastically disable the persistent-conditioning pathway during training, enabling the same model to predict both with and without access to this internal context.
This strategy give rise to a meaningful sampling guidance called Persistence Guidance (PG).
%
For a noisy input $\mathbf{x}_t$, we denote by $\mathbf{x}^{p}_c$ the class-conditional prediction with P-to-A conditioning, by $\mathbf{x}^{p}_u$ its class-unconditional counterpart, and by $\mathbf{x}^{\varnothing}_c$ the class-conditional prediction with P-to-A conditioning disabled.
CFG and PG can then be written symmetrically as
\begin{equation}
\mathrm{CFG}: \mathbf{x}^{\mathrm{CFG}} = \mathbf{x}^{p}_u + w_c (\mathbf{x}^{p}_c - \mathbf{x}^{p}_u), \qquad \mathrm{PG}: \mathbf{x}^{\mathrm{PG}} = \mathbf{x}^{\varnothing}_c + w_p (\mathbf{x}^{p}_c - \mathbf{x}^{\varnothing}_c),
\end{equation}
where $w_c$ and $w_p$ control the strengths of CFG and PG, respectively.
In PG, since two predictions $\mathbf{x}^{p}_c$ and $\mathbf{x}^{\varnothing}_c$ differ only in the availability of persistent conditioning, their contrast $(\mathbf{x}^{p}_c - \mathbf{x}^{\varnothing}_c)$ captures how the sample-specific structural context redirects the ongoing refinement.
Given that persistent features predominantly preserve established global visual structure, PG provides a direction for reinforcing consistency with the structure already formed within the model.
%
Setting $w_p=1$ recovers the standard persistent-conditioned prediction, whereas $w_p>1$ strengthens the influence of the persistent structural context on subsequent refinement.
%


PG and CFG can be naturally combined during to further improve the generation quality as  
\begin{equation}
    \mathbf{x}^{\mathrm{CFG+PG}}
    =
    \mathbf{x}^{p}_u
    +
    w_c (\mathbf{x}^{p}_c - \mathbf{x}^{p}_u)
    +
    (w_p-1) (\mathbf{x}^{p}_c - \mathbf{x}^{\varnothing}_c).
\end{equation}
Intuitively, PG strengthens the influence of the global structural context carried by persistent features during denoising.
Unlike CFG, which relies on external class conditioning, PG derives its guidance direction from internally formed persistent representation, making it a form of internally evolved structural guidance.
Because the two guidance signals arise from different sources of conditioning, they capture distinct yet related aspects of generation and can be naturally complementary.
As illustrated in Fig.~\ref{fig:pg_cfg_analysis}, CFG primarily strengthens class-consistent semantic appearance, whereas PG promotes coherent global organization.
Combining them allows semantic refinement to proceed while remaining anchored to the structural context already established within the model.

\begin{table}[t!]
\centering
\small
\setlength{\tabcolsep}{5pt}
\renewcommand{\arraystretch}{1.1}

\begin{tabularx}{\textwidth}{
l
*{6}{>{\centering\arraybackslash}X}
}
\toprule
\textbf{Model} &
\textbf{Params.} &
\textbf{Epochs} &
\textbf{FID} $\downarrow$ &
\textbf{IS} $\uparrow$ &
\textbf{Prec.} $\uparrow$ &
\textbf{Rec.} $\uparrow$ \\
\midrule

\multicolumn{7}{l}{\textit{\textbf{Latent-space Diffusion Models}}} \\
\midrule

DiT-XL/2 \citep{peebles2023scalable}
& 675M
& 1400
& 2.27
& 278.2
& \textbf{0.83}
& 0.57 \\

SiT-XL/2 \citep{ma2024sit}
& 675M
& 1400
& 2.06
& 277.5
& \textbf{0.83}
& 0.59 \\

REPA-XL/2 \citep{yu2025representation}
& 675M
& 800
& 1.42
& 305.7
& 0.80
& 0.64 \\

DDT-XL/2 \citep{wang2026ddt}
& 675M
& 400
& 1.26
& \textbf{310.6}
& 0.79
& 0.65 \\

RAE-XL/2 \citep{zheng2026diffusion}
& 839M
& 800
& 1.13
& 262.6
& 0.78
& \textbf{0.67} \\

Internal Guidance \citep{zhou2026guiding}
& 678M
& 680
& \textbf{1.07}
& 274.1
& 0.79
& 0.66 \\

\midrule
\multicolumn{7}{l}{\textit{\textbf{Pixel-space Diffusion Models}}} \\
\midrule

ADM-G \citep{dhariwal2021diffusion}
& 554M
& 400
& 4.59
& 186.7
& \textbf{0.82}
& 0.52 \\

RIN \citep{jabri2022scalable}
& 410M
& 480
& 3.42
& 182.0
& --
& -- \\

SiD \citep{hoogeboom2023simple}
& 2B
& --
& 2.44
& 256.3
& --
& -- \\

VDM++ \citep{kingma2023understanding}
& 2B
& --
& 2.12
& 267.7
& --
& -- \\

PixelFlow-XL/4 \citep{chen2025pixelflow}
& 677M
& 320
& 1.98
& 282.1
& \underline{0.81}
& 0.60 \\

PixNerd-XL/16 \citep{wang2026pixnerd}
& 700M
& 320
& 1.93
& 297.0
& 0.79
& 0.59 \\

EPG-G/16 \citep{lei2026there}
& 1391M
& 600
& 1.75
& 275.1
& 0.80
& 0.62 \\

PixelDiT-XL \citep{yu2026pixeldit}
& 797M
& 320
& \textbf{1.61}
& 292.7
& 0.78
& \underline{0.64} \\

DiP-XL/16 \citep{chen2026dip}
& 631M
& 600
& 1.79
& 281.9
& 0.80
& 0.62 \\

DeCo-XL/16 \citep{ma2026deco}
& 682M
& 600
& 1.69
& 304.0
& 0.79
& 0.63 \\

PixelU-H/16 \citep{guo2026pixelu}
& 1168M
& 600
& \underline{1.63}
& 305.9
& 0.79
& \underline{0.64} \\

PixelGen-XL/16 \citep{ma2026pixelgen}
& 676M
& 160
& 1.83
& 293.6
& 0.79
& 0.63 \\

PixelREPA-H/16 \citep{shin2026representation}
& 953M
& 600
& 1.81
& \underline{317.2}
& 0.79
& 0.63 \\

\graymidrule

JiT-B/16~\citep{li2026back}
& 131M
& 600 & 3.66 & 275.1 & \underline{0.81} & 0.52 \\

JiT-L/16~\citep{li2026back}
& 459M
& 600 & 2.36 & 298.5 & 0.79 & 0.60 \\

JiT-H/16~\citep{li2026back}
& 953M
& 600 & 1.86 & 303.4 & 0.78 & 0.62 \\

\graymidrule

\textbf{PerF-B/16}
& 137M
& 600 & 2.81 & 288.3 & \underline{0.81} & 0.56 \\

\textbf{PerF-L/16}
& 471M
& 600 & 1.91 & 311.2 & 0.77 & \underline{0.64} \\

\textbf{PerF-H/16}
& 987M
& 600 & \underline{1.63} & \textbf{324.5} & 0.77 & \textbf{0.65} \\

\bottomrule
\end{tabularx}
\vspace{-5pt}
\caption{
Comparison with latent-space and pixel-space diffusion models on ImageNet $256\times256$.
}
\label{tab:imagenet256}
\end{table}

\begin{table}[ht]
    \centering
    \small
    \setlength{\tabcolsep}{5pt}
    \renewcommand{\arraystretch}{1.1}

    \begin{tabularx}{\textwidth}{
    l
    *{6}{>{\centering\arraybackslash}X}
    }
    \toprule
    \textbf{Model} &
    \textbf{Params.} &
    \textbf{Epochs} &
    \textbf{FID} $\downarrow$ &
    \textbf{IS} $\uparrow$ &
    \textbf{Prec.} $\uparrow$ &
    \textbf{Rec.} $\uparrow$ \\
    \midrule

    \multicolumn{7}{l}{\textit{\textbf{Latent-space Diffusion Models}}} \\
    \midrule

    SiT-XL/2 \citep{ma2024sit}
    & 675M & 600 & 2.62 & 252.2 & \textbf{0.84} & 0.57 \\

    REPA-XL/2 \citep{yu2025representation}
    & 675M & 800 & 2.08 & \textbf{274.6} & 0.83 & 0.58 \\

    RAE-XL/2 \citep{zheng2026diffusion}
    & 839M & 800 & \textbf{1.13} & 259.6 & 0.80 & \textbf{0.63} \\

    \midrule
    \multicolumn{7}{l}{\textit{\textbf{Pixel-space Diffusion Models}}} \\
    \midrule

    ADM-G \citep{dhariwal2021diffusion}
    & 554M & 400 & 7.72 & 172.7 & \textbf{0.84} & 0.53 \\

    PixNerd-XL/16 \citep{wang2026pixnerd}
    & 700M & 340 & 2.84 & 245.6 & 0.80 & 0.59 \\

    EPG-L/32 \citep{lei2026there}
    & 540M & 800 & 2.35 & 295.4 & 0.82 & 0.57 \\

    DeCo-XL/16 \citep{ma2026deco}
    & 682M & 340 & 2.22 & 290.0 & 0.80 & 0.60 \\

    PixelDiT-XL \citep{yu2026pixeldit}
    & 797M & 850 & 1.81 & 278.6 & 0.78 & 0.67 \\

    PixelU-H/32 \citep{guo2026pixelu}
    & 1152M & 600 & 1.92 & 322.1 & 0.80 & 0.58 \\

    \graymidrule

    JiT-H/32~\citep{li2026back}
    & 956M & 600 & 1.94 & 309.1 & 0.80 & 0.61 \\

    \graymidrule

    \textbf{PerF-H/32}
    & 992M & 600 & \textbf{1.76} & \textbf{335.3} & 0.79 & \textbf{0.64} \\

    \bottomrule
    \end{tabularx}
    \vspace{-5pt}
    \caption{
        Comparison with latent-space and pixel-space diffusion models on
        ImageNet $512\times512$.
    }
    \vspace{-10pt}
    \label{tab:imagenet512} 
\end{table}

\section{Experiments}

\subsection{Implementation Details}

We conduct class-conditional generation on ImageNet~\citep{deng2009imagenet} $256\times256$ and $512\times512$ resolutions, and build
PerF upon the JiT architectures~\citep{li2026back}.
%
%
Unless otherwise stated, we follow the JiT optimization and noise-sampling setup.
%
We replace the uniform hidden width with a contraction--expansion schedule, while approximately matching the attention and MLP parameter budget of the corresponding vanilla JiT by preserving $\sum_{\ell=1}^{L} d_\ell^2 = Ld^2$.
Persistent-to-active conditioning is implemented with a low-rank bottleneck of rank $128$ for PerF-B/L and $192$ for PerF-H, keeping the additional parameter overhead below $5\%$ across model scales.
Class conditioning and persistent conditioning are independently dropped with probability $0.1$.
%
We employ REPA~\citep{yu2025representation,singh2026what} on the persistent features with a weight of $0.1$ only for the main experiments.
All ablation and representation analyses are conducted without REPA unless otherwise stated.
More detailed implementation configurations are provided in the Appendix~\ref{sec:app_implementation}.

\subsection{Main Results}

\paragraph{ImageNet $256\times256$.}
Table~\ref{tab:imagenet256} compares PerF with representative latent-space
and pixel-space generative models.
Across all model scales, PerF consistently improves the corresponding JiT
baseline while introducing less than 5\% additional parameters.
PerF-B reduces FID from 3.66 to 2.81, PerF-L from 2.36 to 1.91, and PerF-H
from 1.86 to 1.63, corresponding to relative reductions of approximately
23\%, 19\%, and 12\%, respectively.
The gains therefore persist as the backbone scales from Base to Huge, despite
the increasingly strong JiT baseline.
PerF also consistently improves Inception Score, with PerF-H increasing IS
from 303.4 to 324.5.
Notably, the improvement is accompanied by higher recall across all three
model scales, while precision remains comparable to the corresponding JiT
models.
This suggests that Persistence Forcing improves generative coverage without
simply trading diversity for sample fidelity.
Overall, PerF-H achieves an FID of 1.63 while remaining competitive with the
strongest recent pixel-space models.

\paragraph{ImageNet $512\times512$.}
The same behavior extends to higher-resolution generation in Table~\ref{tab:imagenet512}.
PerF-H/32 improves JiT-H/32 from FID of 1.94 to 1.76  and from IS of 309.1 to 335.3, while recall increases from 0.61 to 0.64.
These gains are obtained without changing the underlying pixel-space generation framework, showing that the benefit of the persistent--active organization transfers across both model scale and image resolution.
Together, the results indicate that Persistence Forcing provides a consistent improvement over its JiT counterpart rather than relying on a particular backbone size or resolution.

\begin{figure}[t]
    \centering
    \includegraphics[width=1.0\linewidth]{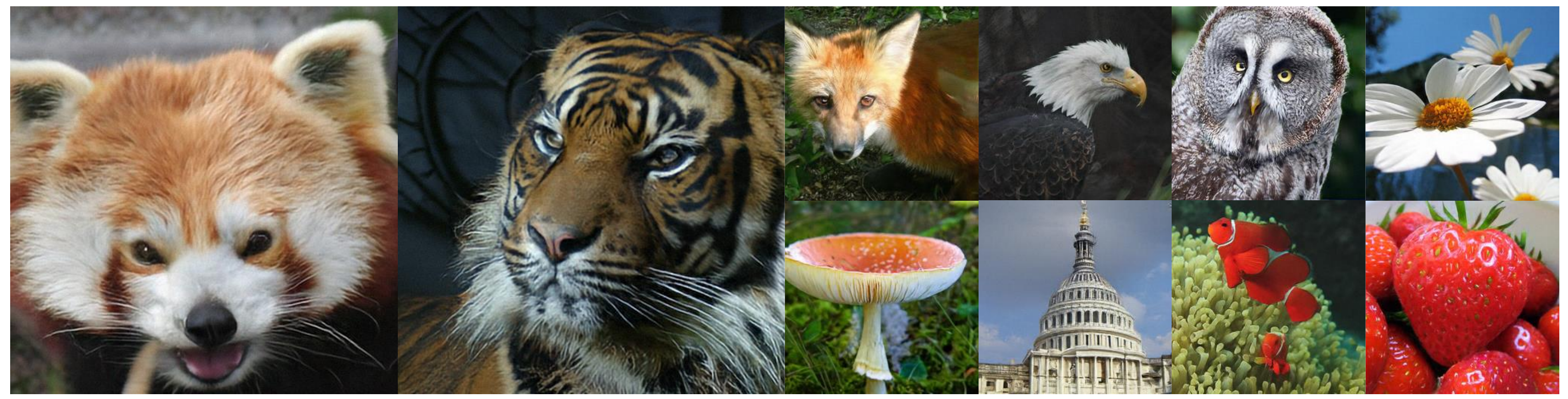}
    \vspace{-15pt}
    \captionof{figure}{
        \textbf{Qualitative results} on ImageNet $512\times512$ and $256\times256$ using PerF-H.
    }
    \vspace{-10pt}
    \label{fig:visual_512_256}
\end{figure}

\subsection{Ablation Study}
We conduct ablation studies on ImageNet $256\times256$ using PerF-B.
Unless otherwise stated, each model variant is trained for 100 epochs without REPA, while sampling-only guidance sweeps reuse the trained PerF-B model from the main experiments.

\paragraph{Refinement schedule.}
\begin{wrapfigure}{r}{0.4\textwidth}
    \centering
    \vspace{-15pt}
    \includegraphics[width=1.0\linewidth]{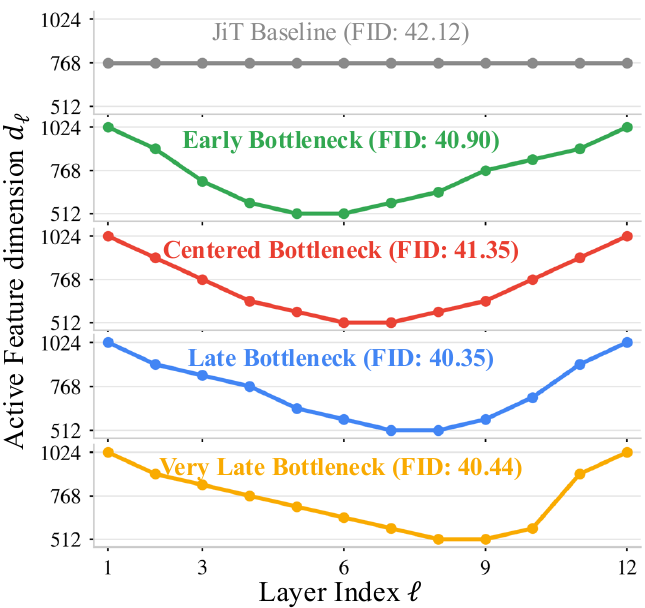}
    \vspace{-20pt}
        \caption{
            \textbf{Ablation on PerF Schedules.}
        }
        \vspace{-10pt}
        \label{fig:ablation_schedule}
    \vspace{-10pt}
\end{wrapfigure}
We study the sensitivity of PerF to the precise contraction--expansion profile by shifting the bottleneck across network depth.
As shown in Figure~\ref{fig:ablation_schedule}, all four schedules achieve similar performance and outperform the uniform JiT baseline at the same training stage, despite substantially different refinement histories.
In particular, FID varies only from $40.35$ to $41.35$ across the tested schedules, compared with $42.1$ for JiT.
This suggests that the effectiveness of PerF does not depend on a narrowly tuned width profile, but is robust to different realizations of heterogeneous refinement.
Among them, schedules with a later bottleneck perform slightly better, with the Late and Very Late variants achieving FIDs of $40.35$ and $40.44$, respectively.
We therefore adopt the Late Bottleneck schedule as our default configuration.

\begin{figure}[t]
    \centering
        \begin{minipage}[t]{0.54\linewidth}
        \vspace{0pt}
        \centering
        \small
        \setlength{\tabcolsep}{1pt}
        \renewcommand{\arraystretch}{1.2}
        \begin{tabular}{
        lc
        cc
        cc
        }
        \toprule
        \multirow{2}{*}{Model} & \multirow{2}{*}{Params.} &
        \multicolumn{2}{c}{w/o CFG} &
        \multicolumn{2}{c}{with CFG} \\
        \cmidrule(lr){3-4}
        \cmidrule(lr){5-6}
        & &
        FID $\downarrow$ &
        IS $\uparrow$ &
        FID $\downarrow$ &
        IS $\uparrow$ \\
        \midrule
        
        PerF (hetero. only)
        & 131M
        & 42.71
        & 36.2
        & 8.02
        & 165.6 \\
        
        $+$ P-to-A Condition
        & 137M
        & 40.35
        & 38.2
        & 7.19
        & 173.8 \\
        
        $+$ Persistence Guidance
        & 137M
        & \textbf{24.92}
        & \textbf{62.1}
        & \textbf{5.81}
        & \textbf{195.8} \\
        
        \bottomrule
        \end{tabular}
        \captionof{table}{\textbf{Progressive construction of PerF.}}
        \label{tab_arch_ablation}
    \end{minipage}
    \hfill
    \begin{minipage}[t]{0.44\textwidth}
        \vspace{0pt}
        \centering
        \small
        \setlength{\tabcolsep}{2pt}
        \begin{tabular}{
        lc
        cc
        cc
        }
        \toprule
        P-to-A & \multirow{2}{*}{Params.} &
        \multicolumn{2}{c}{PG=1.0} &
        \multicolumn{2}{c}{PG=4.0} \\
        \cmidrule(lr){3-4}
        \cmidrule(lr){5-6}
        low-rank $r$ & &
        FID $\downarrow$ &
        IS $\uparrow$ &
        FID $\downarrow$ &
        IS $\uparrow$ \\
        \midrule
        
        32
        & 1.5M
        & 41.66
        & 36.9
        & 28.89
        & 52.9 \\
                    
        64
        & 2.9M
        & 42.24
        & 36.6
        & 26.02
        & 59.1 \\

        \rowcolor{black!8}
        128 (default)
        & 5.7M
        & 40.35
        & 38.2
        & 24.92
        & 62.1\\
        
        256
        & 11.5M
        & 40.06
        & 38.4
        & 22.16
        & 67.2\\
        
        \bottomrule
        \end{tabular}
        \captionof{table}{\textbf{Ablation on P-to-A low-rank $r$.}}
        \label{tab_low_rank}
    \end{minipage}

\end{figure}

\paragraph{Component analysis.}
As established in Table~\ref{tab_vw_only}, heterogeneous refinement alone does not improve generation quality, showing that inducing distinct refinement histories is not sufficient by itself.
Table~\ref{tab_arch_ablation} therefore starts from the heterogeneous backbone and studies how its emergent feature organization is progressively exploited.
Introducing P-to-A conditioning improves FID from 42.71 to 40.35 without CFG and from 8.02 to 7.19 with CFG, demonstrating the benefit of explicitly coupling persistent and actively refined features.
The resulting controllable conditioning pathway further enables Persistence Guidance, which provides the largest additional improvement, reducing FID to 24.92 and 5.81, respectively.
Importantly, PG is not an independent guidance module: its guidance direction is defined by the prediction contrast obtained by enabling and disabling the P-to-A pathway.
The three stages therefore play different roles where heterogeneous refinement induces the feature organization, P-to-A conditioning exposes this organization as an explicit interaction, and PG further exploits its effect during sampling.

\paragraph{P-to-A bottleneck rank.}
Table~\ref{tab_low_rank} studies the capacity of the low-rank projector used for persistent-to-active conditioning.
To isolate the effects of P-to-A conditioning and Persistence Guidance, we disable CFG in this experiment ($\text{CFG}=1$).
Increasing the bottleneck rank generally improves generation quality,
especially when PG is applied.
Without additional PG ($\text{PG}=1$), increasing $r$ from 128 to 256 yields only a marginal FID improvement from 40.35 to 40.06, while doubling the additional parameters from 5.7M to 11.5M.
With stronger PG ($\text{PG}=4$), the benefit of a larger bottleneck becomes more pronounced, suggesting that a higher-capacity conditioning pathway provides a more informative direction for Persistence Guidance.
We use $r=128$ by default as a favorable trade-off between generation quality and parameter overhead.

\begin{figure}[t]
    \centering

        \begin{minipage}[t]{0.64\linewidth}
        \vspace{0pt}
        \centering

        \begin{minipage}[t]{0.56\linewidth}
            \vspace{0pt}
            \centering
            \includegraphics[
                width=\linewidth
            ]{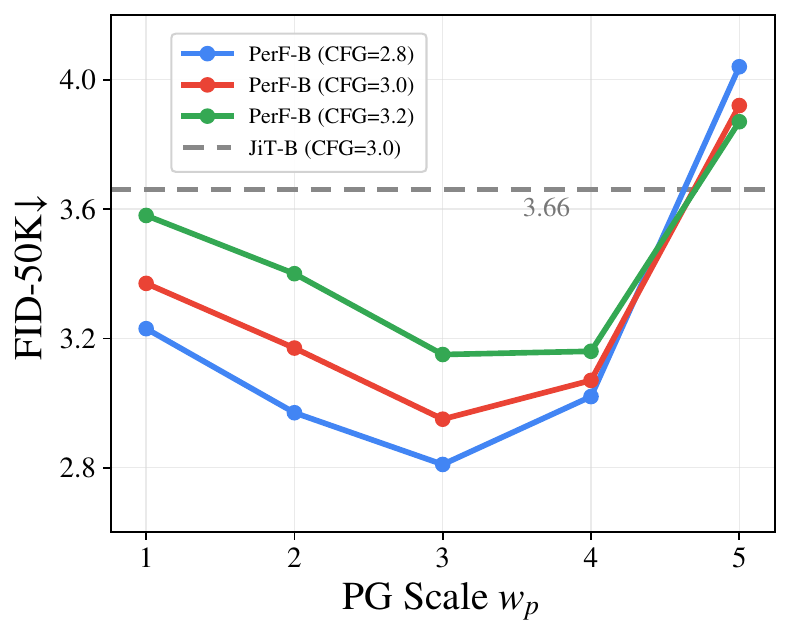}
        \end{minipage}
        \hfill
        \begin{minipage}[t]{0.41\linewidth}
            \vspace{0pt}
            \centering
            \includegraphics[
                width=\linewidth
            ]{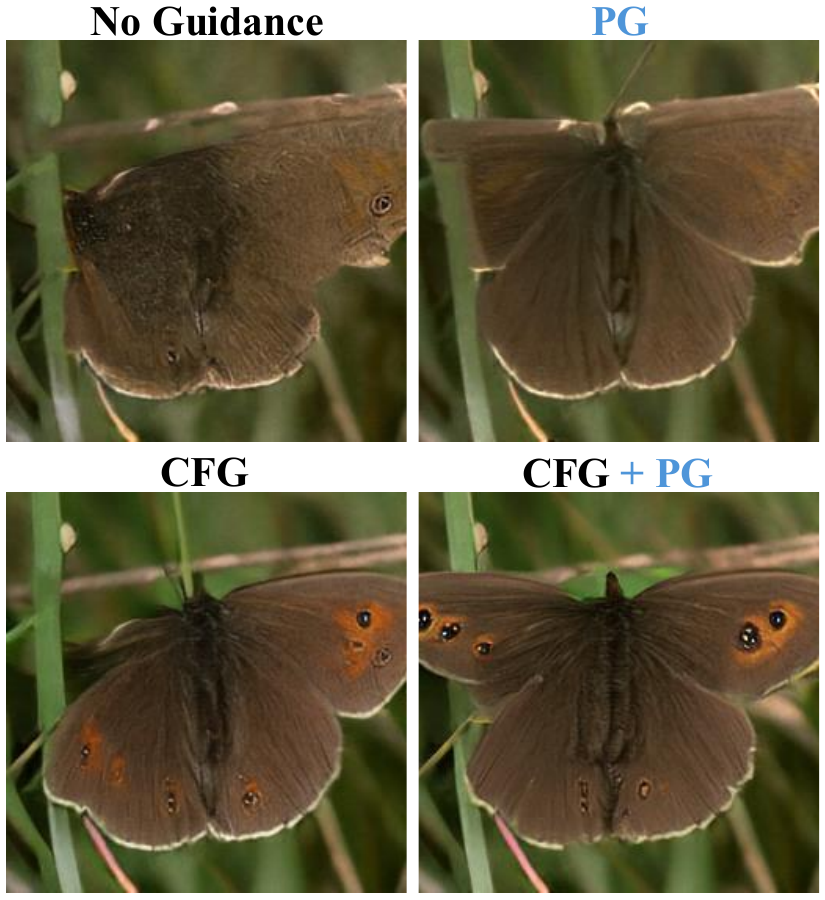}
        \end{minipage}
        \vspace{-10pt}
        \caption{
            \textbf{PG plays a complementary role to CFG.} 
            PG primarily reinforces the global structural organization, whereas CFG strengthens semantic appearance. Combining the two preserves both effects.
        }
        \label{fig:pg_cfg_analysis}
    \end{minipage}
    \hfill
    \begin{minipage}[t]{0.34\linewidth}
        \vspace{3pt}
        \centering
        \small
        \setlength{\tabcolsep}{1pt}
        \renewcommand{\arraystretch}{1.2}

        \begin{tabular}{c cc cc}
            \toprule
            \multirow{2}{*}{PG Interval} &
            \multicolumn{2}{c}{w/o CFG} &
            \multicolumn{2}{c}{with CFG} \\
            \cmidrule(lr){2-3}
            \cmidrule(lr){4-5}
            &
            FID $\downarrow$ &
            IS $\uparrow$ &
            FID $\downarrow$ &
            IS $\uparrow$ \\
            \midrule
            w/o PG
            & 21.05 & 75.0 & 3.37 & 308.3 \\

            \midrule
            \rowcolor{black!8}
            $\textbf{[0,1]}$
            & \textbf{16.03} & \textbf{90.3} & \textbf{3.12} & \textbf{311.0} \\

            $[0.1,1]$
            & 17.40 & 85.6 & 3.16 & 307.4 \\

            $[0,0.9]$
            & 16.17 & 89.3 & 3.23 & 309.6 \\

            $[0.1,0.9]$
            & 17.60 & 84.7 & 3.19 & 306.2 \\

            \bottomrule
        \end{tabular}
        \vspace{-2pt}
        \captionof{table}{
            \textbf{Analysis of PG interval.} We set CFG=3.0 and PG=2.0 when guidance is enabled.
        }
        \label{tab:pg_interval}
    \end{minipage}
    \vspace{-10pt}
\end{figure}

\paragraph{Interaction between PG and CFG.}
For this sampling-only analysis, we reuse the PerF-B model from the
main experiment and vary only the guidance configuration.
Figure~\ref{fig:pg_cfg_analysis} studies whether the benefit of Persistence Guidance can be explained by simply changing the strength of CFG.
Across all tested CFG scales, a moderate amount of PG consistently improves FID, with the best region appearing around $w_p=3$.
The persistence of this gain after CFG is separately tuned suggests that PG does not merely reproduce stronger semantic guidance.
Instead, it provides an additional direction that remains useful over a broad range of CFG strengths.
This behavior is consistent with the different sources of the two guidance signals.
CFG emphasizes externally specified semantic information, while PG reinforces the internally evolved structural context of the current sample.
Performance degrades again when $w_p$ becomes too large, indicating that the persistent context is most effective when it guides active refinement without over-constraining it.
Table~\ref{tab:pg_interval} further studies the denoising interval over which PG is applied.
Applying PG throughout the full interval $[0,1]$ gives the best overall performance both with and without CFG.
These results suggest that the persistent guidance signal remains useful throughout much of the denoising trajectory, rather than being confined to a narrow stage of generation.

\section{Related Work}

\subsection{Learning Representations in Pixel-Space Diffusion Transformers}

Recent pixel-space diffusion models have highlighted the importance of how visual representations are formed and organized within the denoising backbone.
JiT~\citep{li2026back} shows that a simple Transformer operating on large image patches can directly model pixel-space generation without a pretrained tokenizer.
PixelDiT~\citep{yu2026pixeldit} further separates patch-level global modeling from pixel-level local refinement, while DeCo~\citep{ma2026deco} organizes generation along the frequency dimension.
These works highlight the benefit of assigning different forms of visual information to distinct computational roles.
Another line of work directly improves intermediate diffusion representations.
REPA~\citep{yu2025representation,singh2026what} aligns latent diffusion features with pretrained visual representations, and subsequent methods further adapt representation alignment to pixel-space diffusion~\citep{shin2026representation}.
Together, these studies show that both the quality and organization of intermediate representations can substantially influence generative learning.
These perspectives motivate us to study how distinct representational roles can emerge from heterogeneous computation itself.
We find that heterogeneous refinement naturally induces a persistent--active feature specialization, which Persistence Forcing exploits to support ongoing active refinement with preserved persistent representations.

\subsection{Guidance for Diffusion Models}

Sampling guidance is widely used to improve the quality and controllability of
diffusion models.
Classifier-free guidance (CFG)~\citep{ho2022classifier} amplifies the contrast between conditional and unconditional predictions, while AutoGuidance~\citep{karras2024guiding} generalizes this principle by contrasting models of different quality.
A growing line of training-free methods constructs guidance signals by perturbing internal attention or computation paths to obtain alternative predictions for contrastive extrapolation ~\citep{hong2023improving,hong2024smoothed,ahn2024self, hyung2025spatiotemporal,chen2026stochastic}.
Recent Internal Guidance (IG)~\citep{zhou2026guiding} further explores guidance signals arising directly from the internal representations of diffusion Transformers, showing that the internal representational hierarchy can itself provide useful sampling signals. 
Our work derives its signal from another form of internal organization induced by heterogeneous refinement.
Persistent-to-active conditioning exposes a controllable contrast between predictions with and without persistent structural context, which PG amplifies during sampling.
The resulting signal reinforces internally evolved structural information and complements the external semantic conditioning provided by CFG.

\section{Conclusion}
In this work, we revisit the common assumption of uniform refinement in pixel-space diffusion Transformers and show that heterogeneous refinement gives rise to an ordered form of feature specialization.
Features receiving limited refinement predominantly preserve coherent global visual structure, while more frequently refined features increasingly focus on finer, high-frequency details, revealing a natural persistent--active organization within the denoising backbone.
Building on this observation, we introduce \emph{Persistence Forcing (PerF)}, which explicitly leverages persistent representations as structural context for active refinement and further amplifies their influence through sampling-time Persistence Guidance.
Across model scales and image resolutions, PerF consistently improves the corresponding JiT baselines, achieving FIDs of $1.63$ and $1.76$ with the Huge models on ImageNet $256\times256$ and $512\times512$, respectively.
These results suggest that organizing computation around emergent feature roles, rather than refining all hidden representations uniformly, provides a promising direction for pixel-space generative modeling and motivates further study of heterogeneous refinement in broader diffusion architectures and applications.

\bibliography{iclr2027_conference}
\bibliographystyle{iclr2027_conference}

\appendix

\section{Extended Related Work}

\paragraph{Pixel-Space Diffusion Models}
Pixel-space diffusion models directly model images without relying on a pretrained tokenizer~\citep{rombach2022high,yao2025reconstruction,fu2026improving,zheng2026diffusion}.
Early works demonstrate the feasibility of diffusion directly in pixel space~\citep{dhariwal2021diffusion,jabri2022scalable,hoogeboom2023simple,kingma2023understanding,wang2025reconciling}, while recent approaches increasingly adopt Transformer-based architectures.
PixelFlow~\citep{chen2025pixelflow} and JiT~\citep{li2026back} show that scalable Transformer backbones can effectively operate on raw pixels, with JiT further demonstrating that large image patches provide a simple and strong design for pixel-space generation.
Subsequent works improve pixel modeling through different forms of architectural organization. 
PixNerd~\citep{wang2026pixnerd} models intra-patch details with neural fields, while DiP~\citep{chen2026dip} and PixelDiT~\citep{yu2026pixeldit} separate global structure modeling from local detail refinement.
DeCo~\citep{ma2026deco} instead decouples generation along the frequency dimension, and PixelU~\citep{guo2026pixelu} introduces a U-shaped Transformer to combine semantic abstraction with fine-grained spatial information.
Other works improve pixel-space generation through self-supervised pretraining, perceptual objectives, or by exploiting the internal representations of frozen pretrained models for self-guidance~\citep{lei2026there,ma2026pixelgen,fu2026frozen}.
In contrast to these explicitly designed spatial, frequency, or objective decompositions, our work studies how representational roles can emerge from
heterogeneous computation itself.
By assigning different feature groups heterogeneous refinement budgets, PerF induces a persistent--active organization within a single Transformer backbone and subsequently exploits this emergent specialization for generation.

\paragraph{Nonuniform Computation in Transformers}
Nonuniform computation has been explored extensively as a way to allocate model capacity according to computational demand.
Early approaches vary network depth or width through adaptive execution, layer skipping, or slimmable architectures~\citep{graves2016adaptive, wang2018skipnet,yu2019universally}.
In Transformers, related methods distribute computation nonuniformly across layers or tokens through structured layer dropping, token pruning, and dynamic routing~\citep{rao2021dynamicvit,raposo2024mixture}.
More recently, Variable-Width Transformers~\citep{wu2026variable} explicitly vary hidden width across network depth, demonstrating that nonuniform layer-wise capacity allocation can alter computational efficiency in decoder-only large language models.
Nonuniform computation has also been introduced into diffusion Transformers. DyDiT~\citep{zhao2025dynamic} dynamically adapts model width across diffusion timesteps and selectively processes spatial tokens, while subsequent methods vary token density or computation across layers, spatial regions, and denoising stages~\citep{chang2025sparsedit}.
Rather than using nonuniform computation primarily for efficiency or capacity allocation, PerF uses heterogeneous refinement as a mechanism for inducing and exploiting structured specialization within the hidden representation.

\paragraph{Hierarchical Organization in Pixel-Space Diffusion}
Hierarchical organization has emerged as a recurring principle in generative modeling, where global or low-frequency structure is established before finer visual details are resolved.
Recent pixel-space models explicitly instantiate this principle through architectural decomposition.
DiP~\citep{chen2026dip} separates global structure construction from local detail restoration, while PixelDiT~\citep{yu2026pixeldit} employs patch-level and pixel-level Transformers for global semantic modeling and local texture refinement, respectively.
Related hierarchical designs progressively vary spatial resolution or patch granularity across network depth~\citep{dao2026multi}.
Other methods impose coarse-to-fine organization along the frequency or denoising trajectory.
DeCo~\citep{ma2026deco} separates low-frequency semantic modeling from high-frequency detail generation, while Latent Forcing~\citep{baade2026latent} encourages semantic latent representations to mature before high-frequency pixel information.
Spectral Forcing~\citep{fan2026show} instead makes the time-dependent spectral structure of pixel-space diffusion explicit by exposing signal-bearing frequency components while suppressing noise-dominated frequencies.
These approaches explicitly prescribe a spatial, frequency, or temporal ordering for generation.
In contrast, PerF does not assign coarse and fine roles a priori: heterogeneous refinement causes such an ordering to emerge across feature groups with different refinement histories.

\section{Analysis on Feature Specialization}

\begin{figure}[t]
    \centering
    \includegraphics[width=1.0\linewidth]{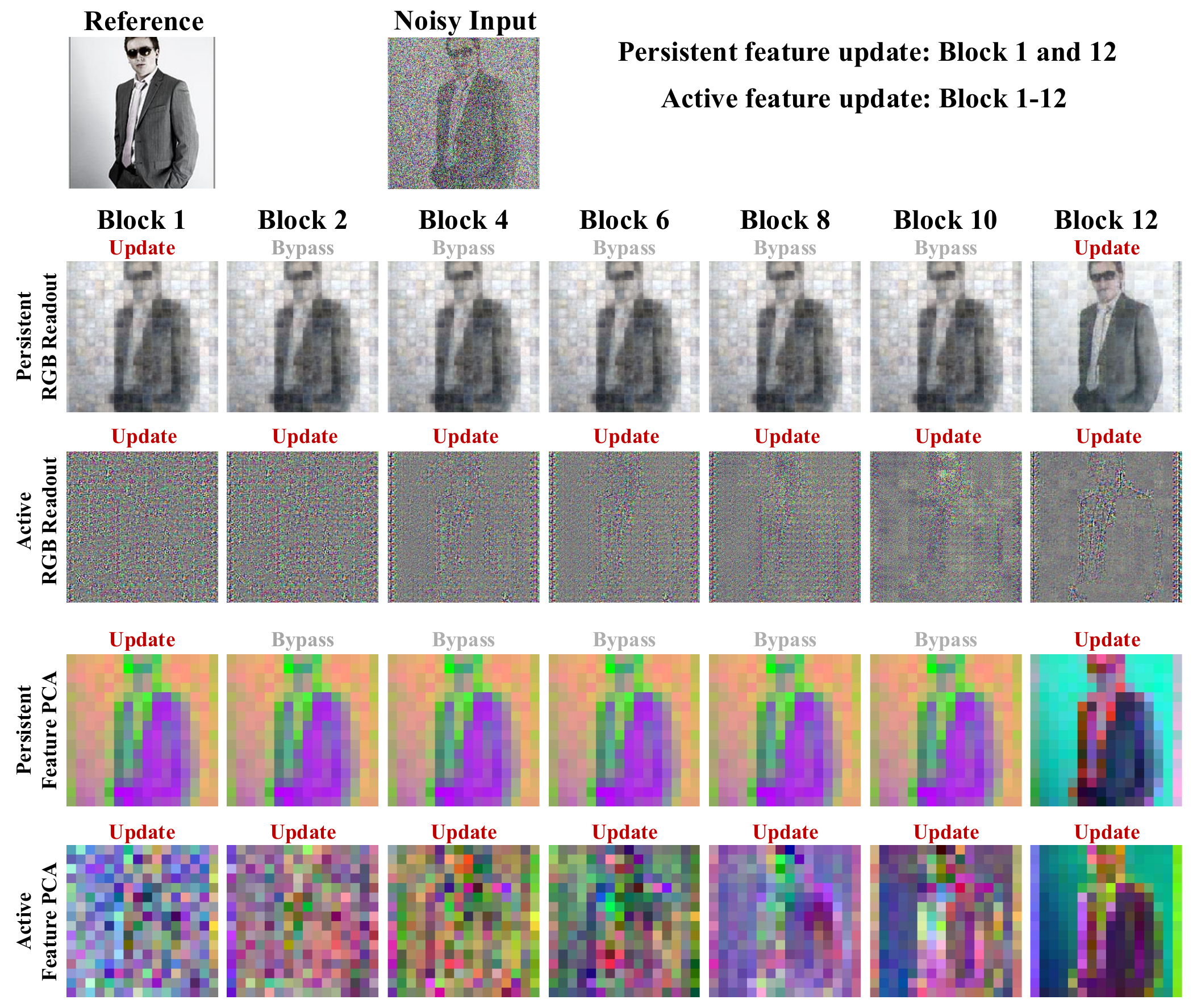}
    \caption{
    \textbf{Depth-wise evolution of persistent and active features.} We visualize the intermediate RGB readouts and feature-space PCA representations of a persistent feature group and the corresponding actively refined features. The persistent features are updated at Blocks 1 and 12 and bypass the intermediate blocks, whereas the active features are updated at every block. After only the first update, persistent features already exhibit coherent global object structure, which remains stable throughout the bypass trajectory before being further refined upon re-entry at Block 12. In contrast, active features continue to evolve across depth and predominantly exhibit residual and higher-frequency structures at intermediate blocks.
    }
    \label{fig:feature_trajectory}
\end{figure}

\subsection{Depth-wise Evolution of Persistent and Active Features}
\label{sec:feature_trajectory}

Our previous analyses characterize the different roles of persistent and active features through their contributions to the final prediction.
We further examine how these representations evolve \emph{within} the network, particularly whether persistent features already contain structured information before their final re-entry.
Figure~\ref{fig:feature_trajectory} visualizes the depth-wise trajectories of a representative persistent feature group and the corresponding actively refined features.
In this example, the persistent group is updated only at Blocks 1 and 12, while bypassing the intermediate blocks, whereas the active features are updated throughout the network.
For a controlled comparison, we select 128-dimensional persistent and active feature groups, ensuring that both visualizations use the same number of channels.

Remarkably, after only the first update, the persistent features already exhibit a coherent representation of the global object structure.
Both their RGB readout and feature-space PCA visualization remain highly stable throughout the subsequent bypass trajectory from Blocks 2 to 10.
This observation is important because the structural organization is already present \emph{before} the persistent features re-enter computation at the final block.
It therefore cannot be explained solely as a property acquired after late feature mixing.
After re-entry at Block 12, the persistent representation is further refined and incorporates more detailed visual information.

The actively refined features exhibit a markedly different trajectory. 
Their intermediate RGB readouts remain dominated by residual and high-frequency patterns, while their feature-space organization continuously changes across depth.
Only toward later blocks do increasingly recognizable local structures and boundaries emerge.
Together, these trajectories provide direct evidence that heterogeneous refinement induces distinct depth-wise representational behaviors: persistent features acquire globally organized information after limited refinement and preserve it during bypass, whereas active features continue to evolve as computation proceeds.
This behavior further motivates the use of persistent features as a stable
conditioning signal for ongoing active refinement in
Section~\ref{sec:persistent_conditioning}.

\subsection{Quantifying Feature Specialization in the Frequency Domain}
\label{sec:spectral_analysis}

\begin{figure}[t]
    \centering
    \includegraphics[width=0.6\linewidth]{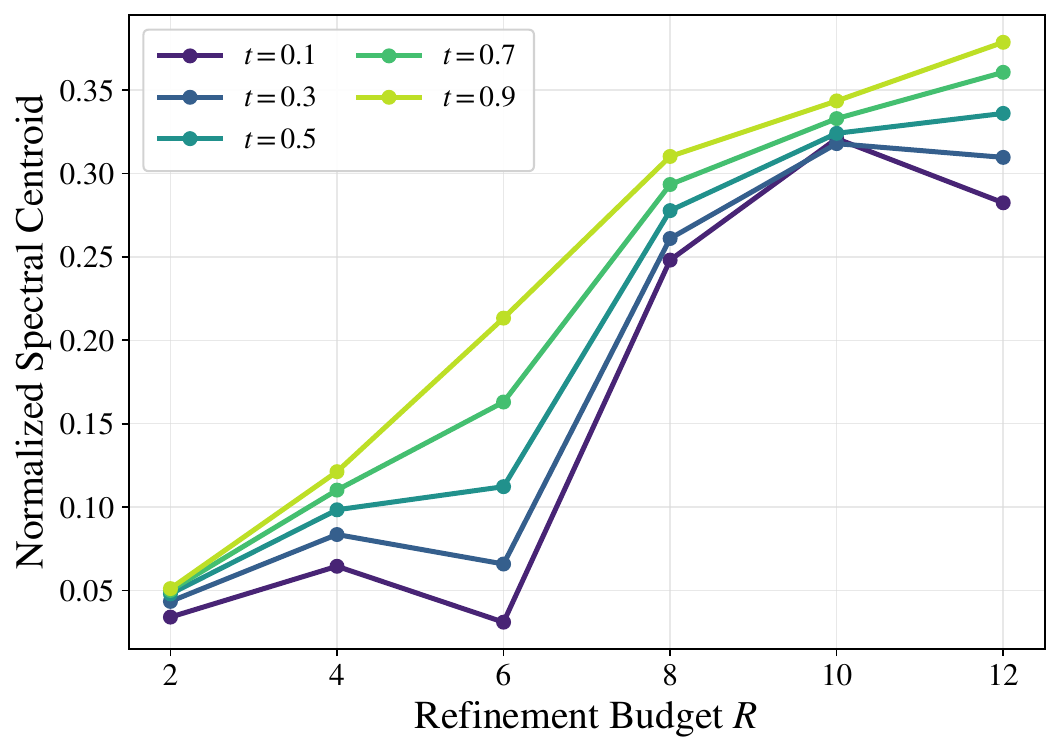}
    \caption{
    \textbf{Frequency specialization across refinement budgets.}
    We measure the normalized spectral centroid of the RGB contribution from
    feature groups with different refinement budgets $R$, using equal-sized
    64-channel subsets for a controlled comparison.
    Larger refinement budgets generally correspond to higher spectral centroids
    across diffusion timesteps, quantitatively confirming the coarse-to-fine
    feature organization observed in Figure~\ref{fig:feature_specialization}.
    }
    \label{fig:spectral_centroid}
\end{figure}

The qualitative decomposition in Figure~\ref{fig:feature_specialization} suggests that feature
groups with different refinement budgets develop systematically different
spatial characteristics.
We further quantify this specialization in the frequency domain by measuring
the spectral content of each group's contribution to the final RGB prediction.

\paragraph{Frequency metric.}
For each feature group, we first obtain its additive RGB contribution through
the model's output projection.
To avoid confounding refinement budget with the number of contributing
channels, we evaluate each group using an equal-sized subset of $64$ feature
dimensions.
We subtract the spatial mean and compute the 2D Fourier
power spectrum of the RGB contribution.
We then measure its normalized spectral centroid, defined as the
energy-weighted mean radial spatial frequency.
Higher values therefore indicate that a larger fraction of the contribution is
concentrated toward finer spatial frequencies.

\paragraph{Frequency organization across refinement budgets.}
Figure~\ref{fig:spectral_centroid} reports the normalized spectral centroid
across refinement budgets at several diffusion timesteps.
A clear coarse-to-fine organization emerges.
Feature groups with small refinement budgets ($R=2$--$6$) are consistently
dominated by low spatial frequencies, whereas groups receiving more refinement
exhibit substantially higher spectral centroids.
The largest transition occurs between intermediate and highly refined groups,
with $R\geq8$ contributing increasingly finer spatial structure.
Importantly, the same overall trend is observed across all evaluated
timesteps, indicating that the relationship between refinement budget and
frequency specialization is not restricted to a particular point along the
denoising trajectory.
Although the ordering is not strictly monotonic for every individual
timestep, larger refinement budgets consistently shift the representation
toward higher-frequency content overall.

These results quantitatively support the qualitative observation in
Figure~\ref{fig:feature_specialization}: heterogeneous refinement organizes feature groups
according to their refinement histories, with sparsely refined features
primarily carrying coarse global structure and more actively refined features
progressively specializing toward finer visual details.

\begin{table*}[t]
\centering
\scriptsize
\setlength{\tabcolsep}{4pt}
\renewcommand{\arraystretch}{1.15}

\begin{tabularx}{\textwidth}{
@{}
l
c
c
c
>{\raggedright\arraybackslash}X
@{}
}
\toprule
\textbf{Model} &
\textbf{Depth} &
$\boldsymbol{d_{\max}}$ &
$\boldsymbol{d_{\min}}$ &
\textbf{Width schedule $\{d_\ell\}$} \\
\midrule

PerF-B/16 &
12 &
1024 &
512 &
[1024, 896, 832, 768, 640, 576, 512, 512, 576, 704, 896, 1024]
\\[2pt]

PerF-L/16 &
24 &
1600 &
512 &
[1600, 1472, 1408, 1280, 1152, 1088, 1088, 960, 896, 832, 832, 768,
704, 640, 640, 576, 576, 512, 640, 768, 896, 1088, 1280, 1600]
\\[2pt]

PerF-H/16 &
32 &
2000 &
640 &
[2000, 1920, 1840, 1760, 1600, 1520, 1440, 1360,
1360, 1280, 1200, 1120, 1040, 1040, 1040, 960,
880, 800, 800, 800, 720, 720, 720, 640,
800, 880, 1040, 1120, 1280, 1520, 1680, 2000]
\\

\bottomrule
\end{tabularx}
\caption{\textbf{Layer-wise width schedules used in PerF.}
For each model scale, the variable-width backbone is approximately
parameter-matched to the corresponding JiT backbone by preserving the
attention and MLP parameter budget,
$\sum_{\ell=1}^{L} d_\ell^2 \approx Ld^2$.
}
\label{tab:width_schedule}
\end{table*}

\begin{table*}[t]
\centering
\small
\setlength{\tabcolsep}{6pt}
\renewcommand{\arraystretch}{1.08}

\begin{tabular}{lcccc}
\toprule
&
\textbf{PerF-B/16} &
\textbf{PerF-L/16} &
\textbf{PerF-H/16} &
\textbf{PerF-H/32} \\
\midrule

\multicolumn{5}{l}{\textbf{Architecture}} \\
\midrule
Image size
    & 256 & 256 & 256 & 512 \\
Depth
    & 12 & 24 & 32 & 32 \\
Patch size
    & 16 & 16 & 16 & 32 \\
Hidden dim $d_{\max}$
    & 1024 & 1600 & 2000 & 2000 \\
Hidden dim $d_{\min}$
    & 512 & 512 & 640 & 640 \\
P-to-A rank
    & 128 & 128 & 192 & 192 \\
Parameters
    & 137M & 471M & 987M & 992M \\
\midrule

\multicolumn{5}{l}{\textbf{Training}} \\
\midrule
Epochs
    & 600 & 600 & 600 & 600 \\
Optimizer
    & \multicolumn{4}{c}{Adam, $\beta_1=0.9,\ \beta_2=0.95$} \\
Batch size
    & \multicolumn{4}{c}{1024} \\
Learning rate
    & \multicolumn{4}{c}{2e-4} \\
Learning rate schedule
    & \multicolumn{4}{c}{constant} \\
Weight decay
    & \multicolumn{4}{c}{0} \\
EMA decay
    & \multicolumn{4}{c}{$\{0.9996, 0.9999\}$} \\
Time sampler
    & \multicolumn{4}{c}{
    $\operatorname{logit}(t)\sim\mathcal{N}(-0.8,0.8^2)$} \\
Noise scale
    & 1.0 & 1.0 & 1.0 & 2.0 \\
Class drop prob.
    & \multicolumn{4}{c}{0.1} \\
P-to-A cond. drop prob.
    & \multicolumn{4}{c}{0.1} \\
\midrule

\multicolumn{5}{l}{\textbf{Sampling}} \\
\midrule
ODE solver
    & \multicolumn{4}{c}{Heun} \\
ODE steps
    & \multicolumn{4}{c}{50} \\
CFG scale
    & 2.8 & 2.7 & 3.0 & 2.9 \\
PG scale
    & 3.0 & 2.8 & 1.9 & 2.0 \\
CFG interval
    & \multicolumn{4}{c}{$[0.1,1.0]$} \\
PG interval
    & \multicolumn{4}{c}{$[0,1]$} \\
\bottomrule
\end{tabular}

\caption{
\textbf{Configurations of PerF experiments.}
}
\label{tab:implementation_details}
\end{table*}

\section{Implementation Details}\label{sec:app_implementation}

Table~\ref{tab:implementation_details} summarizes the architecture, training, and sampling
configurations of our main PerF models.
We provide additional implementation details below.

\paragraph{Architecture details.}
We build PerF-B, PerF-L, and PerF-H upon the corresponding JiT
architectures~\citep{li2026back}.
For each model scale, we replace the uniform hidden width with a fixed
contraction--expansion schedule, where the $\ell$-th Transformer block
actively transforms only the first $d_\ell$ feature dimensions and the
remaining dimensions bypass the block unchanged.
The complete layer-wise width schedules are reported in
Table~\ref{tab:width_schedule}.
To isolate the effect of heterogeneous refinement from changes in backbone
capacity, each variable-width backbone approximately matches the attention
and MLP parameter budget of its corresponding JiT model by preserving
\[
\sum_{\ell=1}^{L} d_\ell^2 \approx Ld^2,
\]
where $d$ denotes the hidden dimension of the corresponding uniform-width
JiT backbone.
The P-to-A conditioning pathway introduces only a small additional parameter
overhead.
We use a low-rank bottleneck of rank $r=128$ for PerF-B/L and $r=192$ for
PerF-H.
PerF-H/16 and PerF-H/32 share the same Transformer depth, hidden dimensions,
and layer-wise width schedule.
Class conditioning and P-to-A conditioning are independently dropped with
probability $0.1$ during training.

\paragraph{Training details.}
Unless otherwise specified, we follow the optimization and noise-sampling
setup of JiT~\citep{li2026back}.
All main models are trained for 600 epochs with Adam
($\beta_1=0.9$, $\beta_2=0.95$), using a global batch size of 1024, a
constant learning rate of $2\times10^{-4}$, and no weight decay.
The diffusion timestep is sampled according to
$\operatorname{logit}(t)\sim\mathcal{N}(-0.8,0.8^2)$.
We track exponential moving averages with decay rates
$\{0.9996,0.9999\}$.
The noise scale is set to $1.0$ for $256\times256$ generation and $2.0$
for $512\times512$ generation.

For the main scaling results in Tables~\ref{tab:imagenet256}
and~\ref{tab:imagenet512}, we additionally employ
REPA~\citep{yu2025representation,singh2026what} during only the first
100 training epochs.
Extending representation alignment throughout the full training schedule
degrades final generation quality, consistent with similar observations
in~\citep{shin2026representation,wang2025repa}.
All ablation variants that require retraining are trained without REPA,
unless otherwise stated.
Sampling-only PG analyses instead reuse the corresponding trained checkpoint
from the main experiments and vary only the sampling configuration.

\paragraph{Sampling details.}
We use the Heun solver with 50 sampling steps for all reported results.
CFG is applied over the interval $[0.1,1.0]$, while PG is applied over the
full interval $[0,1]$ unless otherwise specified.
For PerF-B/16, PerF-L/16, PerF-H/16, and PerF-H/32, we use CFG scales of
$2.8$, $2.7$, $3.0$, and $2.9$, respectively, with corresponding PG scales
of $3.0$, $2.8$, $1.9$, and $2.0$.
The detailed configurations for each model are summarized in
Table~\ref{tab:implementation_details}.

\section{Implementation of Persistent-to-Active Conditioning}
The P-to-A projector is instantiated and learned independently at each Transformer block.
At block $\ell$, the hidden representation is partitioned into active features
$\mathbf{h}^{a}_{\ell}\in\mathbb{R}^{N\times d_{\ell}}$ and persistent
features
$\mathbf{h}^{p}_{\ell}\in\mathbb{R}^{N\times (D-d_{\ell})}$.
Persistent features are first normalized and then mapped through a low-rank
two-layer projector,
\[
\mathbf{m}^{p}_{\ell}
=
W^{(2)}_{\ell}
\,\mathrm{SiLU}\!\left(
W^{(1)}_{\ell}\,
\mathrm{RMSNorm}(\mathbf{h}^{p}_{\ell})
\right),
\]
where
\[
W^{(1)}_{\ell}\in
\mathbb{R}^{r\times(D-d_{\ell})},
\qquad
W^{(2)}_{\ell}\in
\mathbb{R}^{6d_{\ell}\times r}.
\]
Thus, the bottleneck rank $r$ directly controls the capacity of the
persistent-conditioning pathway.
The resulting token-wise modulation
$\mathbf{m}^{p}_{\ell}\in\mathbb{R}^{N\times6d_{\ell}}$
is split into six components corresponding to the shift, scale, and gate
parameters of the self-attention and MLP sublayers.
These terms are added to the standard timestep- and class-conditioned AdaLN
modulation~\citep{wang2026ddt} before updating the active features.
The persistent features themselves bypass the block unchanged and are
concatenated back with the updated active representation afterward.

During training, the entire P-to-A pathway is independently dropped for each
sample with probability $0.1$, enabling the same model to produce predictions
with and without persistent conditioning for Persistence Guidance.
The final linear layer of each P-to-A projector is zero-initialized so that
the persistent-conditioning branch is introduced as a zero residual at the
start of training.

\section{Generation without Classifier-free Guidance.}
Table~\ref{tab:imagenet256_nocfg} evaluates PerF without external
classifier-free guidance.
The improvement over the corresponding JiT baselines remains consistent
across model scales, showing that the benefit of PerF does not rely on CFG or
its interaction with the class-conditioning direction.
Instead, the persistent-conditioning pathway and its induced guidance remain
effective on their own, supporting the view that PerF provides an internal
signal complementary to external semantic guidance.
The simultaneous improvements in IS and recall further suggest that this
benefit extends beyond a simple shift in sample fidelity.

\begin{table}[ht]
\centering
\small
\setlength{\tabcolsep}{2.0pt}
\renewcommand{\arraystretch}{1.12}

\begin{tabular}{
@{}
l c c c c c c
@{}
}
\toprule
\textbf{Model} &
\textbf{Params.} &
\textbf{Epochs} &
\textbf{FID} $\downarrow$ &
\textbf{IS} $\uparrow$ &
\textbf{Prec.} $\uparrow$ &
\textbf{Rec.} $\uparrow$ \\

\midrule

ADM~\citep{dhariwal2021diffusion}
& 554M & 400 & 10.94 & -- & 0.69 & 0.63 \\

PixelFlow-XL~\citep{chen2025pixelflow}
& 677M & 320 & 12.23 & 103.3 & 0.63 & 0.66 \\

PixNerd-XL~\citep{wang2026pixnerd}
& 700M & 320 & 15.61 & 88.9 & 0.59 & 0.68 \\

DeCo-XL/16~\citep{ma2026deco}
& 682M & 320 & 14.88 & 88.2 & 0.60 & 0.68 \\

PixelGen-XL/16~\citep{ma2026pixelgen}
& 676M & 80 & 5.11 & 159.2 & \textbf{0.72} & 0.63 \\

JiT-B/16~\citep{li2026back} 
& 131M & 600 & 25.42 & 63.2 & 0.54 & 0.66 \\

JiT-L/16~\citep{li2026back} 
& 458M & 600 & 13.85 & 104.2 & 0.62 & 0.67 \\

JiT-H/16~\citep{li2026back} 
& 953M & 600 &  7.15 & 151.7 & 0.68 & 0.67 \\

\midrule

\textbf{PerF-B/16} 
& 137M 
& 600  
& 12.53
& 104.0
& 0.63 
& 0.66
\\

\textbf{PerF-L/16} 
& 471M 
& 600  
& 5.60
& 160.7
& 0.68 
& \textbf{0.69}
\\

\textbf{PerF-H/16} 
& 987M 
& 600  
& \textbf{3.48}
& \textbf{194.4}
& 0.70
& \textbf{0.69}
\\

\bottomrule
\end{tabular}

\caption{
Comparison of class-conditional generation on ImageNet at
$256\times256$ resolution without CFG.
All other settings and notation follow Table~\ref{tab:imagenet256}.
}
\label{tab:imagenet256_nocfg}
\end{table}

\begin{table}[ht]
\centering
\begin{minipage}[t]{0.48\linewidth}
\centering
\small
\begin{tabular}{lc}
\toprule
\textbf{Model / Training Recipe} & \textbf{FID} $\downarrow$ \\
\midrule
JiT-B/16 & 3.66 \\
PerF-B/16 w/o REPA & 3.10 \\
PerF-B/16 + REPA (full) & 3.38 \\
PerF-B/16 + REPA (first 100 ep.) & \textbf{2.81} \\
\bottomrule
\end{tabular}
\caption*{\textbf{(a) REPA schedule in PerF-B/16.}}
\end{minipage}
\hfill
\begin{minipage}[t]{0.48\linewidth}
\centering
\small
\begin{tabular}{lcc}
\toprule
\textbf{Model} & \textbf{FID} $\downarrow$ & \textbf{IS} $\uparrow$ \\
\midrule
PixelREPA-B/16 & 3.17 & 284.6 \\
PerF-B/16 & \textbf{2.81} & \textbf{288.3} \\
\midrule
PixelREPA-L/16 & 2.11 & 309.5 \\
PerF-L/16 & \textbf{1.91} & \textbf{311.2} \\
\midrule
PixelREPA-H/16 & 1.81 & 317.2 \\
PerF-H/16 & \textbf{1.63} & \textbf{324.5} \\
\bottomrule
\end{tabular}
\caption*{\textbf{(b) Comparison with PixelREPA.}}
\end{minipage}

\caption{\textbf{Interaction with representation alignment.}
Left: effect of different REPA schedules on PerF-B/16.
Right: comparison with PixelREPA, which specifically adapts representation
alignment to JiT.}
\label{tab:repa_interaction}
\end{table}

\section{Interaction with Representation Alignment.}
We further study how Persistence Forcing interacts with representation
alignment.
Different from standard REPA~\citep{yu2025representation,singh2026what},
we apply the alignment objective only to the persistent feature subspace,
leaving actively refined features unconstrained to model finer pixel-level
details.

Table~\ref{tab:repa_interaction}(a) shows that PerF already improves over
JiT-B without representation alignment, indicating that its benefit does not
depend on REPA.
Moreover, applying REPA throughout training is less effective than using no
alignment, consistent with recent observations that prolonged representation
alignment can interfere with later-stage generative learning
~\citep{shin2026representation,wang2025repa}.
Restricting alignment to the first 100 epochs instead gives the strongest
result, suggesting that representation alignment is most useful during early
optimization while the persistent representation is still being formed.

We further compare against PixelREPA~\citep{shin2026representation} in
Table~\ref{tab:repa_interaction}(b), which is specifically designed to make
representation alignment effective for JiT.
PerF consistently achieves better FID across B, L, and H scales, while also
improving IS.
Together, these results show that the gains of PerF cannot be explained by
representation alignment alone.
Instead, REPA serves as an auxiliary
early-training objective that further improves the persistent representation
learned by PerF.

\section{Architectural Overhead}
Table~\ref{tab:compute_cost} compares the parameter count and single-forward computational cost of PerF with the corresponding JiT backbones.

\begin{table}[ht]
\centering
\small
\setlength{\tabcolsep}{6pt}
\renewcommand{\arraystretch}{1.15}
\begin{tabular}{lcccccc}
\toprule
\textbf{Scale} &
\multicolumn{2}{c}{\textbf{JiT}} &
\multicolumn{2}{c}{\textbf{PerF}} &
\multicolumn{2}{c}{\textbf{Overhead}} \\
\cmidrule(lr){2-3}
\cmidrule(lr){4-5}
\cmidrule(lr){6-7}
&
\textbf{Params.} &
\textbf{GFLOPs} &
\textbf{Params.} &
\textbf{GFLOPs} &
\textbf{Params.} &
\textbf{GFLOPs} \\
\midrule
Base  & 131M & 25.22  & 137M & 26.51  & +4.6\% & +5.1\% \\
Large & 459M & 87.94  & 471M & 90.83  & +2.6\% & +3.3\% \\
Huge  & 953M & 181.95 & 987M & 190.44 & +3.6\% & +4.7\% \\
\bottomrule
\end{tabular}
\caption{\textbf{Per-forward computational cost on ImageNet
$256\times256$.}
PerF introduces modest parameter and FLOP overhead over the corresponding
JiT backbones across model scales.}
\label{tab:compute_cost}
\end{table}
Despite introducing heterogeneous refinement and persistent-to-active conditioning, PerF incurs only modest overhead across model scales.
The additional parameters remain below $5\%$, while the increase in single-forward FLOPs is approximately $3$--$5\%$.
This shows that the persistent--active interaction can be incorporated with limited architectural overhead.

\section{Additional Qualitative Results}
\begin{figure*}[p]
  \centering

  \begin{minipage}[t]{0.495\textwidth}
  \centering
  \includegraphics[width=\linewidth]{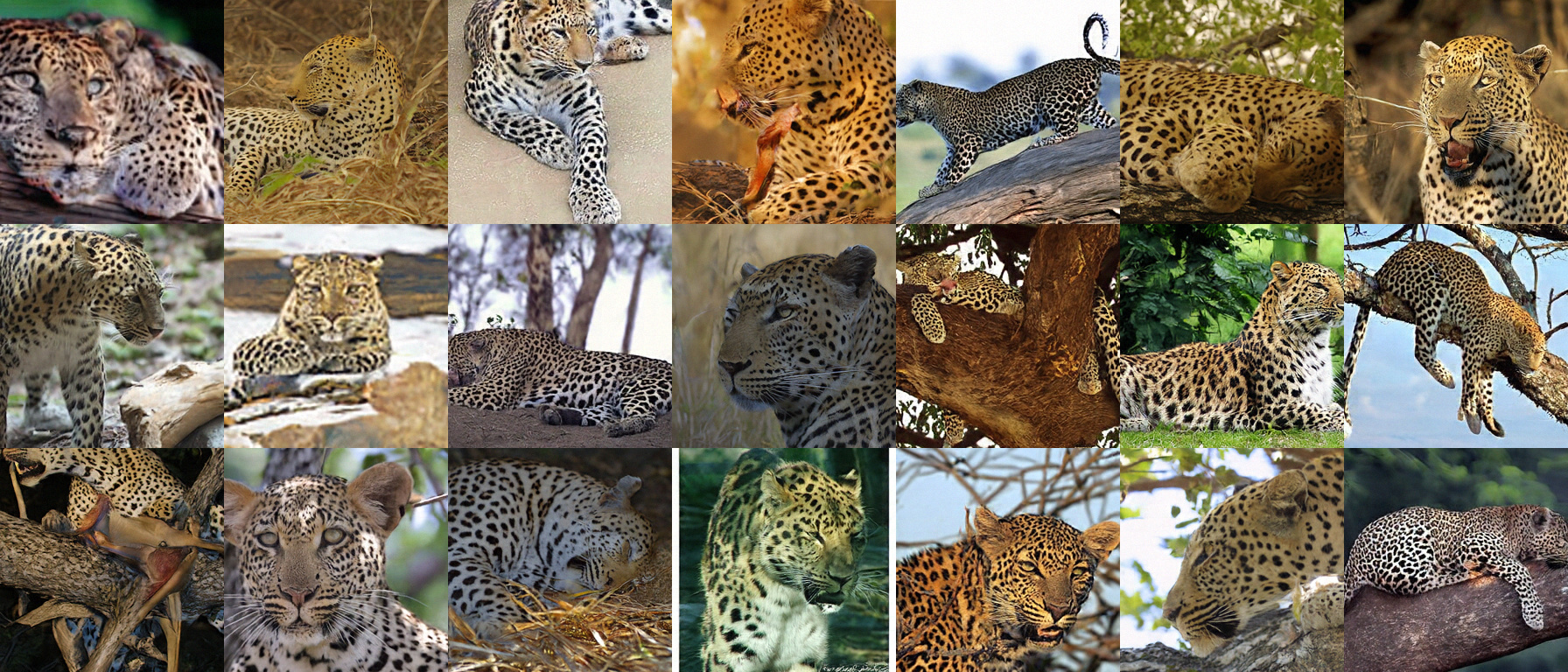}
  \par\vspace{-0.25em}
  {\scriptsize class 288: leopard, Panthera pardus\par}
  \end{minipage}\hfill
  \begin{minipage}[t]{0.495\textwidth}
  \centering
  \includegraphics[width=\linewidth]{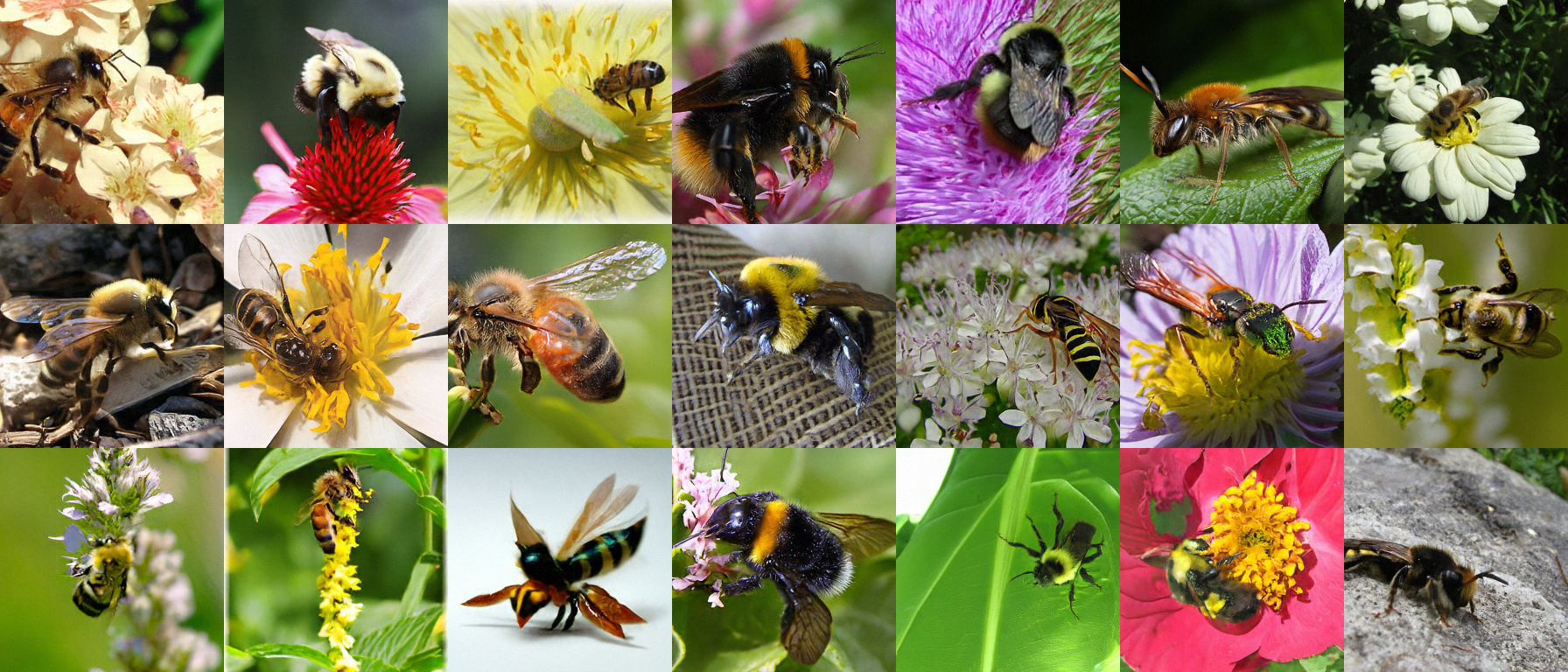}
  \par\vspace{-0.25em}
  {\scriptsize class 309: bee\par}
  \end{minipage}

  \vspace{0.15em}

  \begin{minipage}[t]{0.495\textwidth}
  \centering
  \includegraphics[width=\linewidth]{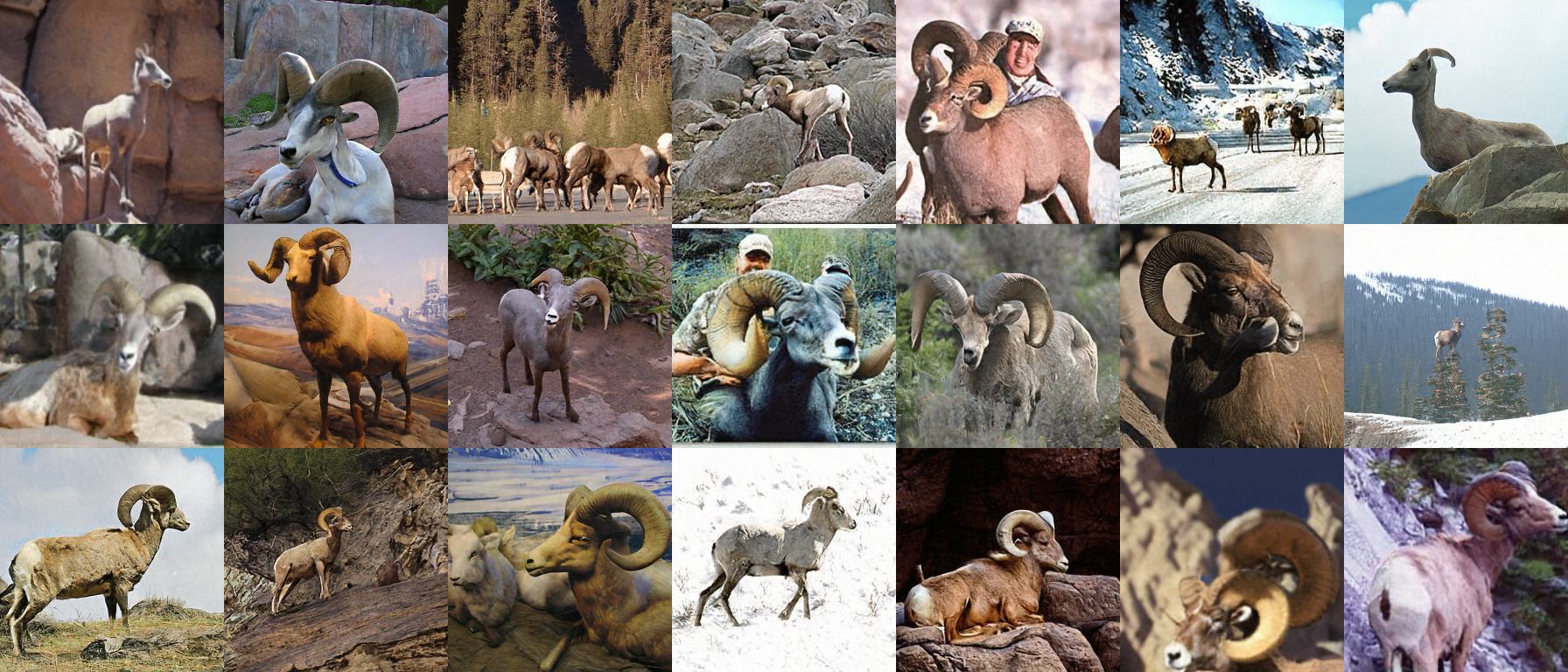}
  \par\vspace{-0.25em}
  {\scriptsize class 349: bighorn, bighorn sheep, cimarron, Rocky Mountain bighorn\par}
  \end{minipage}\hfill
  \begin{minipage}[t]{0.495\textwidth}
  \centering
  \includegraphics[width=\linewidth]{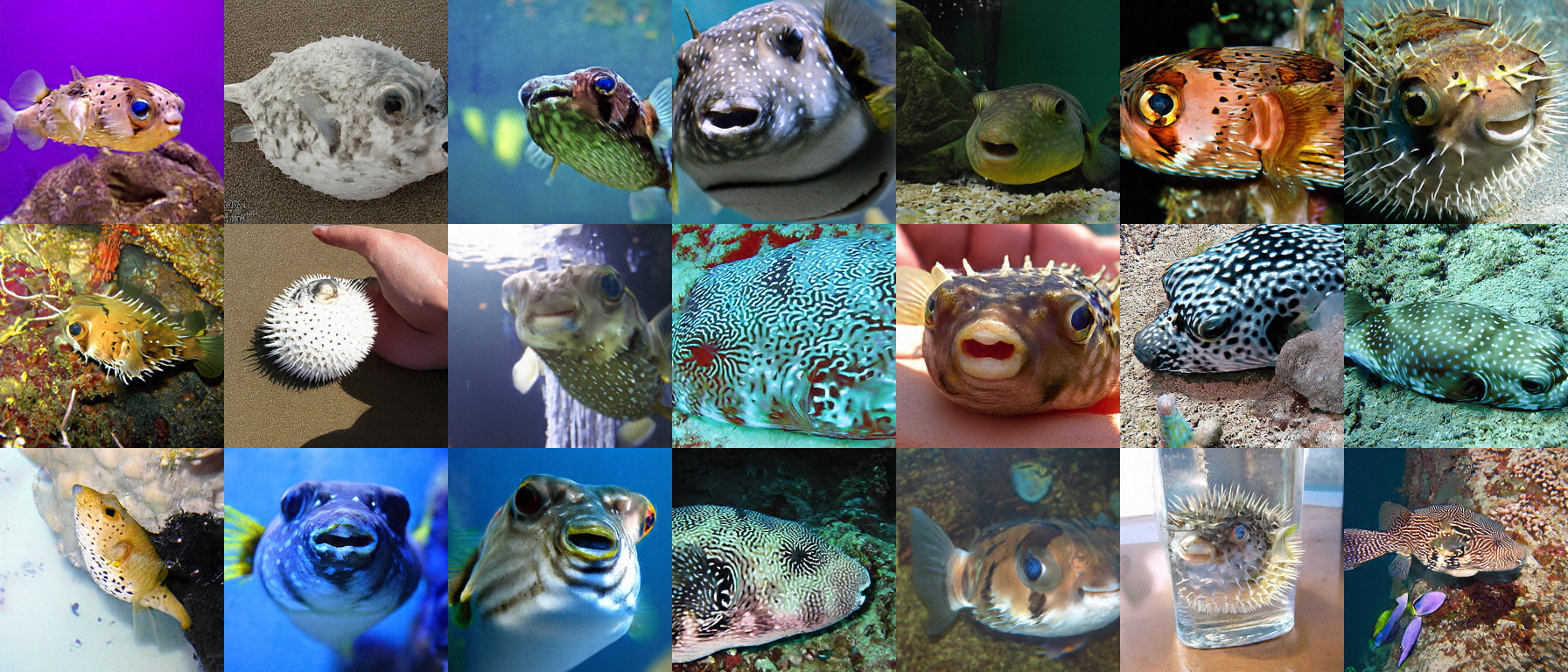}
  \par\vspace{-0.25em}
  {\scriptsize class 397: puffer, pufferfish, blowfish, globefish\par}
  \end{minipage}

  \vspace{0.15em}

  \begin{minipage}[t]{0.495\textwidth}
  \centering
  \includegraphics[width=\linewidth]{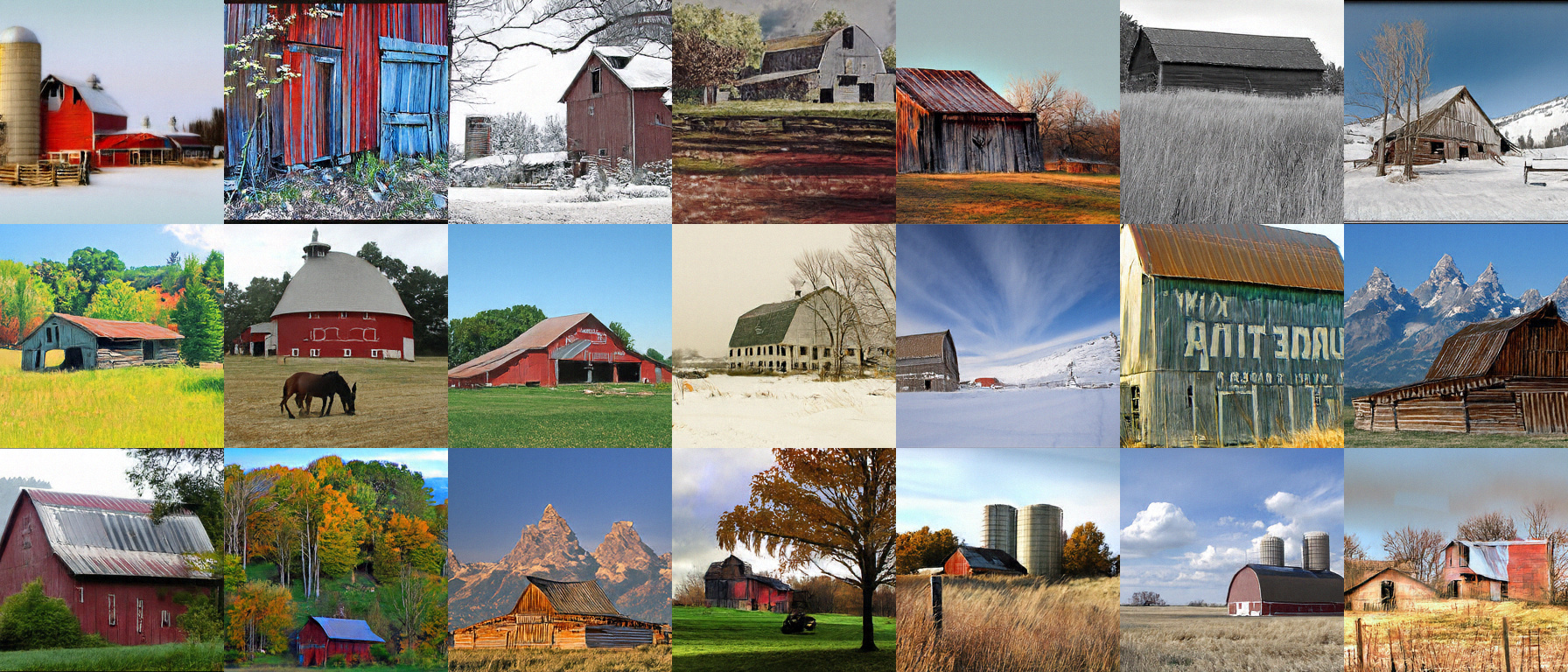}
  \par\vspace{-0.25em}
  {\scriptsize class 425: barn\par}
  \end{minipage}\hfill
  \begin{minipage}[t]{0.495\textwidth}
  \centering
  \includegraphics[width=\linewidth]{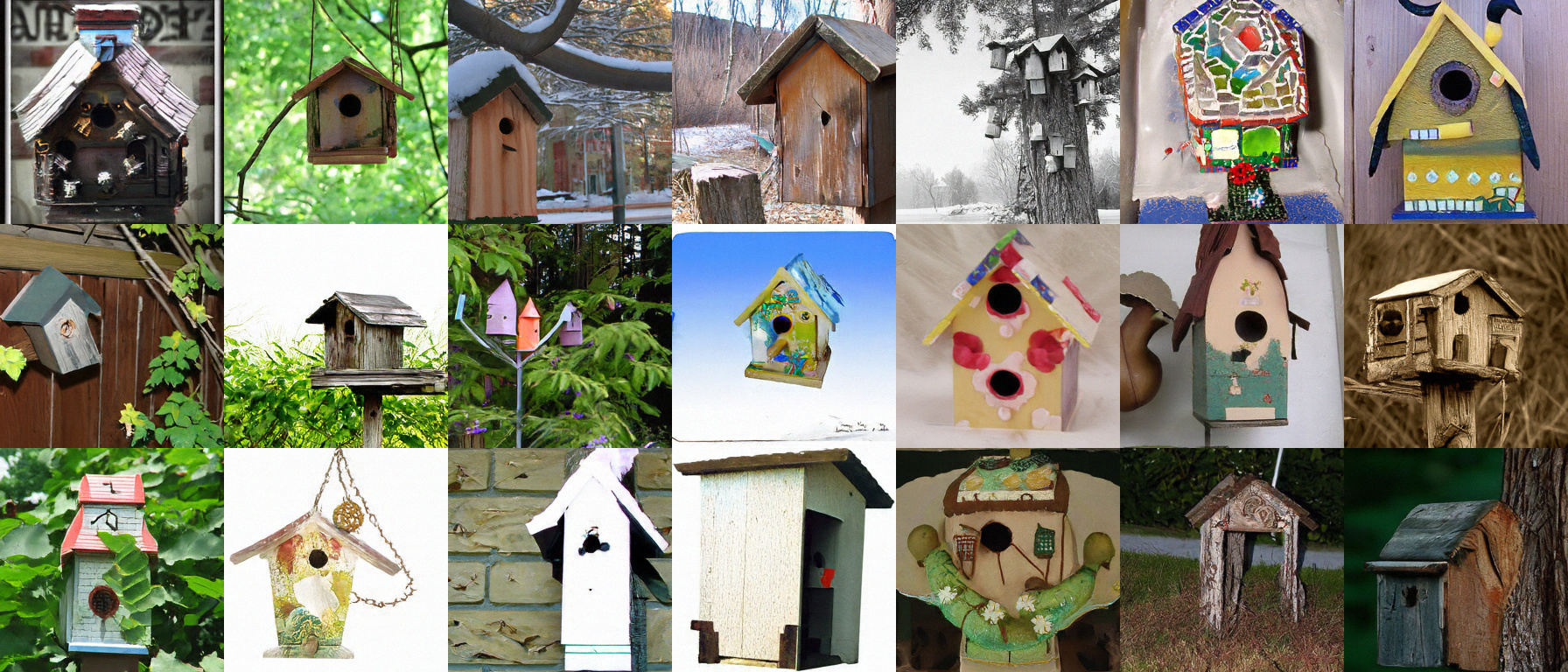}
  \par\vspace{-0.25em}
  {\scriptsize class 448: birdhouse\par}
  \end{minipage}

  \vspace{0.15em}

  \begin{minipage}[t]{0.495\textwidth}
  \centering
  \includegraphics[width=\linewidth]{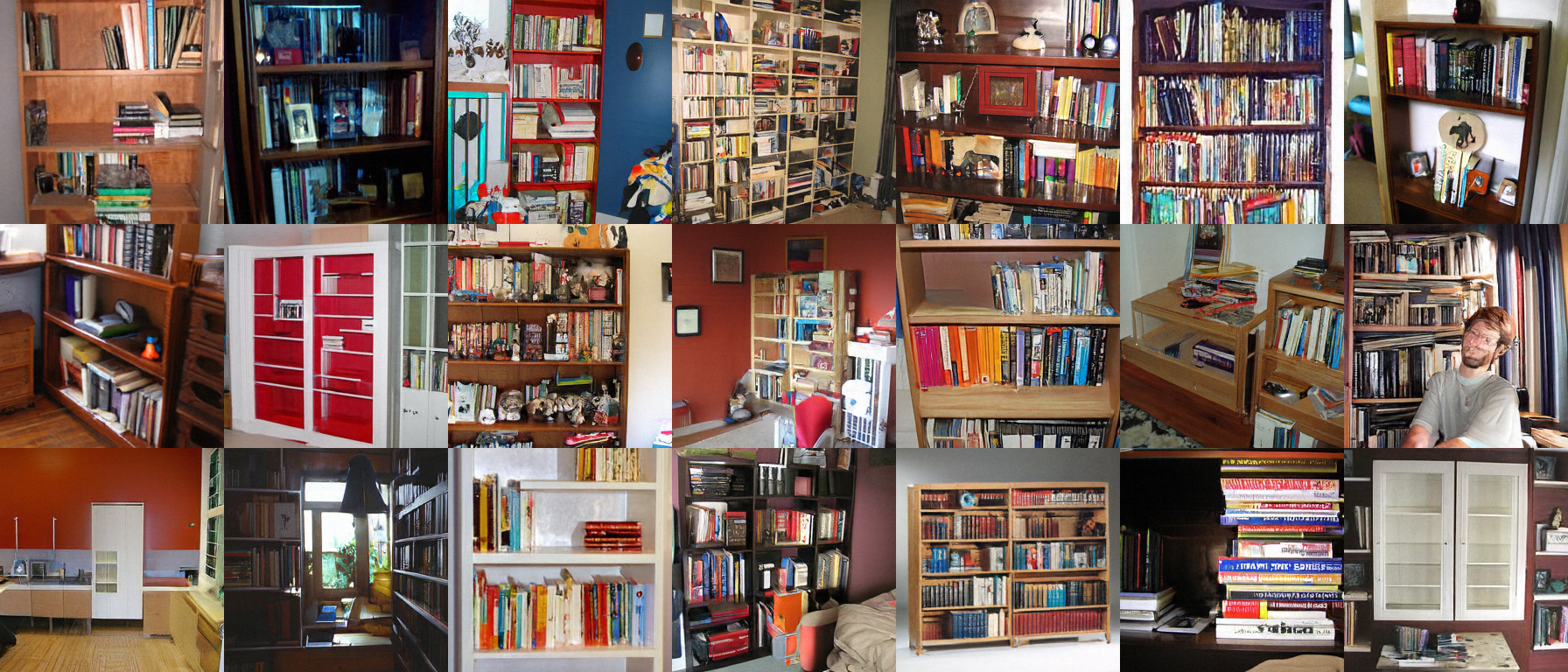}
  \par\vspace{-0.25em}
  {\scriptsize class 453: bookcase\par}
  \end{minipage}\hfill
  \begin{minipage}[t]{0.495\textwidth}
  \centering
  \includegraphics[width=\linewidth]{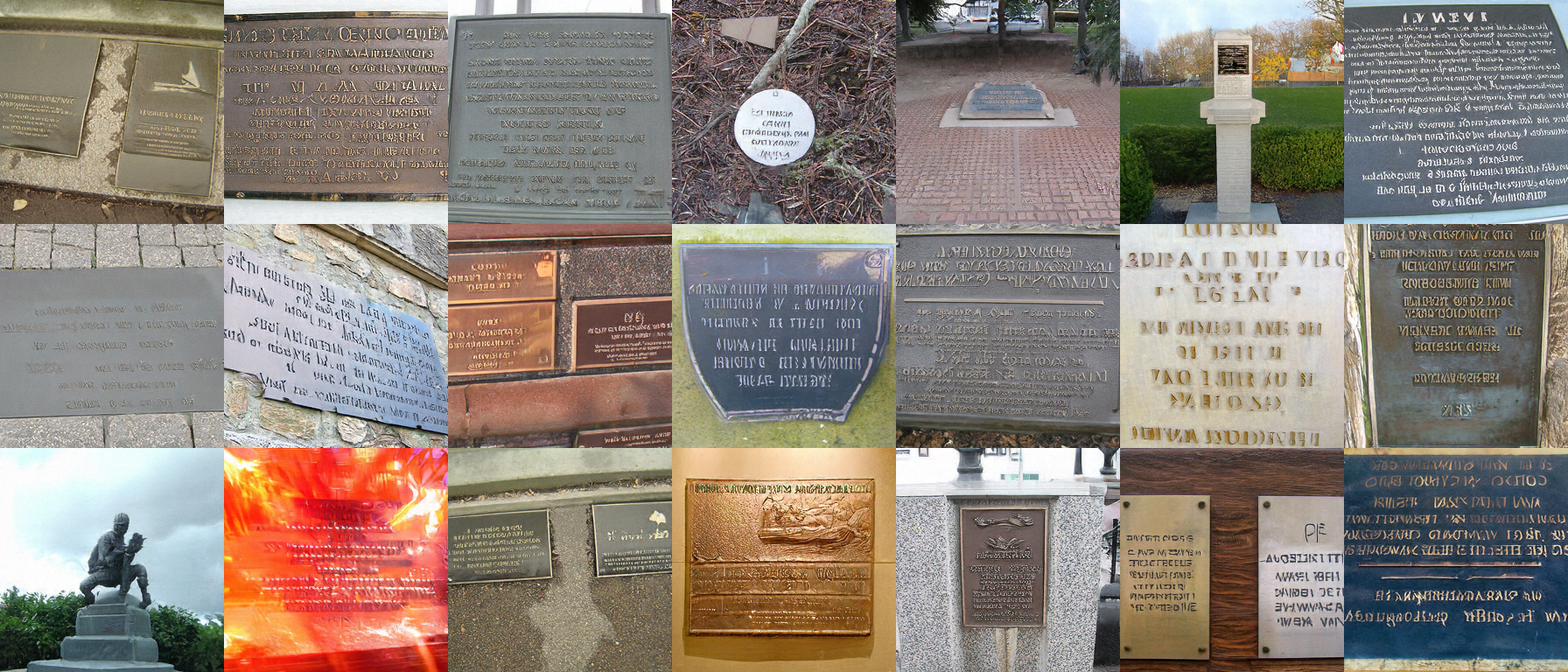}
  \par\vspace{-0.25em}
  {\scriptsize class 458: brass, memorial tablet, plaque\par}
  \end{minipage}

  \vspace{0.15em}

  \begin{minipage}[t]{0.495\textwidth}
  \centering
  \includegraphics[width=\linewidth]{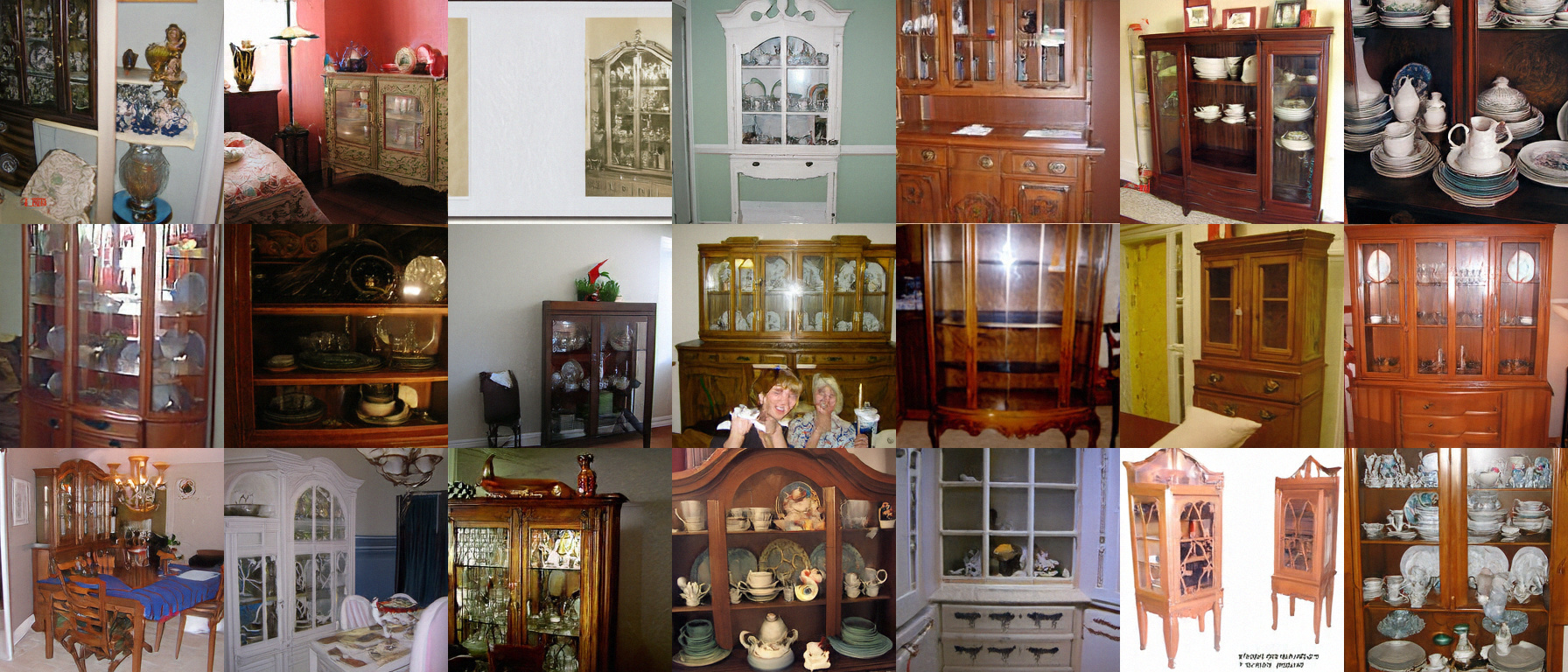}
  \par\vspace{-0.25em}
  {\scriptsize class 495: china cabinet, china closet\par}
  \end{minipage}\hfill
  \begin{minipage}[t]{0.495\textwidth}
  \centering
  \includegraphics[width=\linewidth]{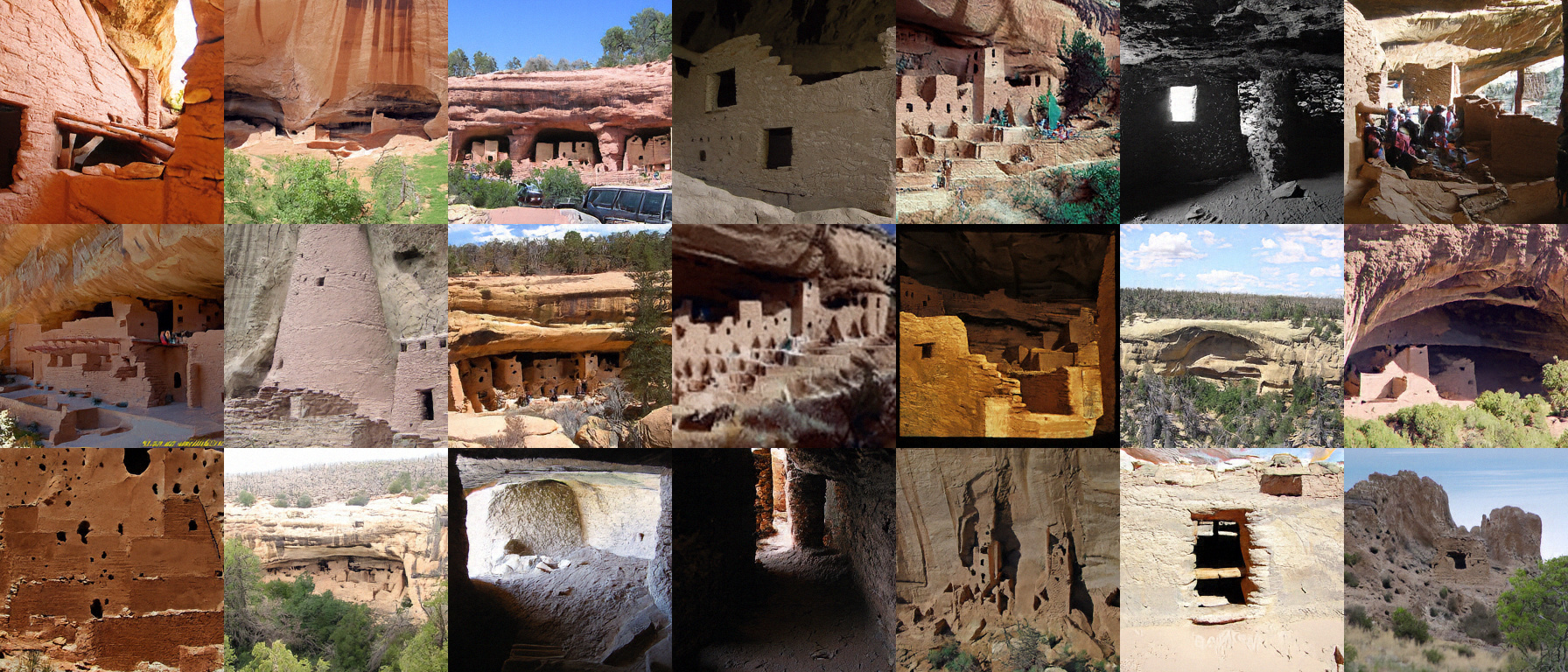}
  \par\vspace{-0.25em}
  {\scriptsize class 500: cliff dwelling\par}
  \end{minipage}

  \caption{Uncurated class-conditional samples on ImageNet $256\times256$ using PerF-H/16}
  \label{fig:perf_h16_samples_1}
\end{figure*}

\begin{figure*}[p]
  \centering

  \begin{minipage}[t]{0.495\textwidth}
  \centering
  \includegraphics[width=\linewidth]{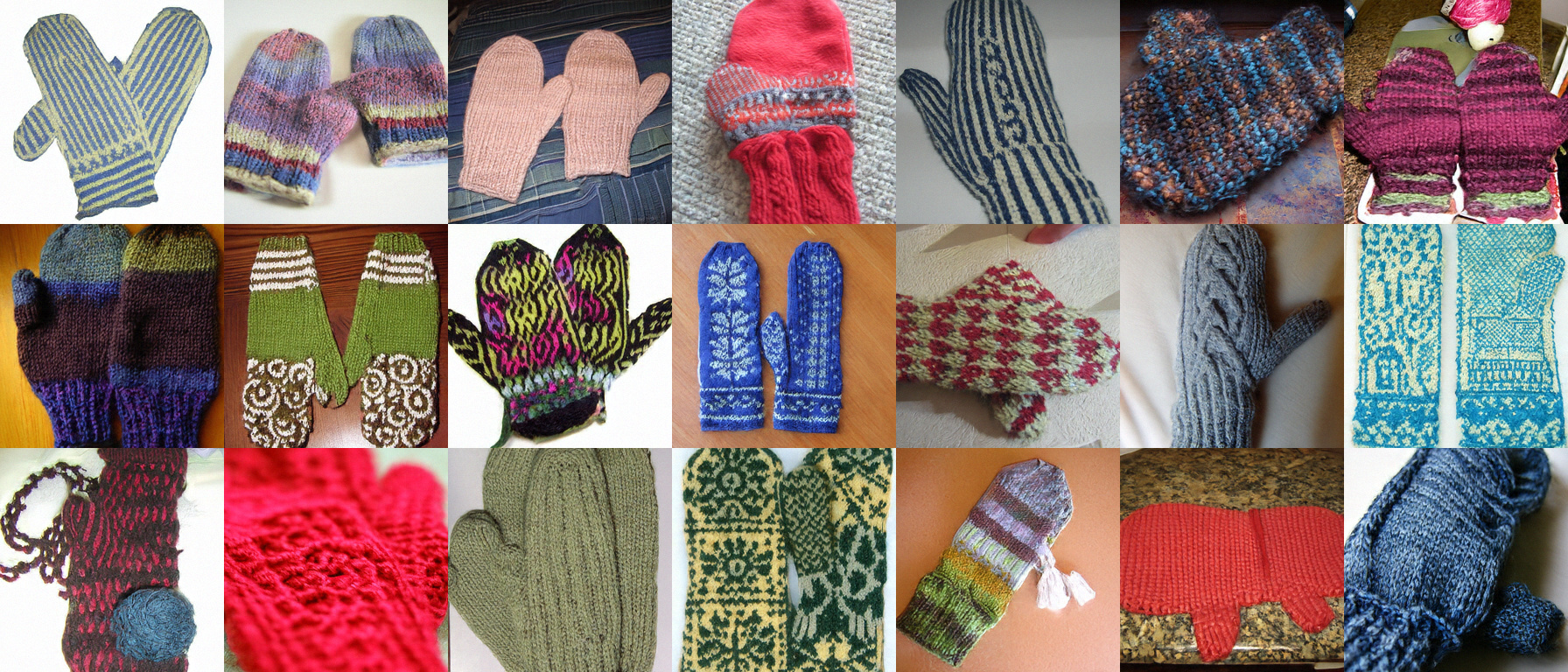}
  \par\vspace{-0.25em}
  {\scriptsize class 658: mitten\par}
  \end{minipage}\hfill
  \begin{minipage}[t]{0.495\textwidth}
  \centering
  \includegraphics[width=\linewidth]{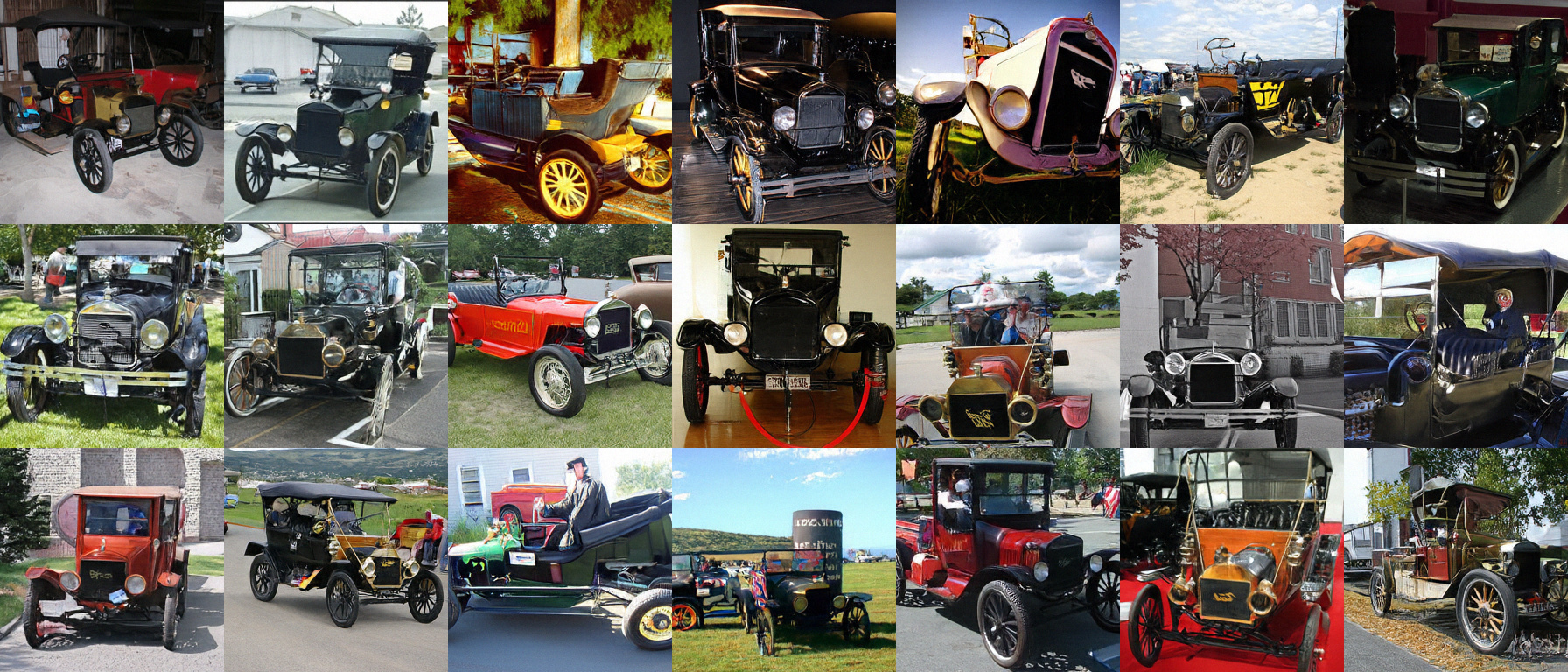}
  \par\vspace{-0.25em}
  {\scriptsize class 661: Model T\par}
  \end{minipage}

  \vspace{0.15em}

  \begin{minipage}[t]{0.495\textwidth}
  \centering
  \includegraphics[width=\linewidth]{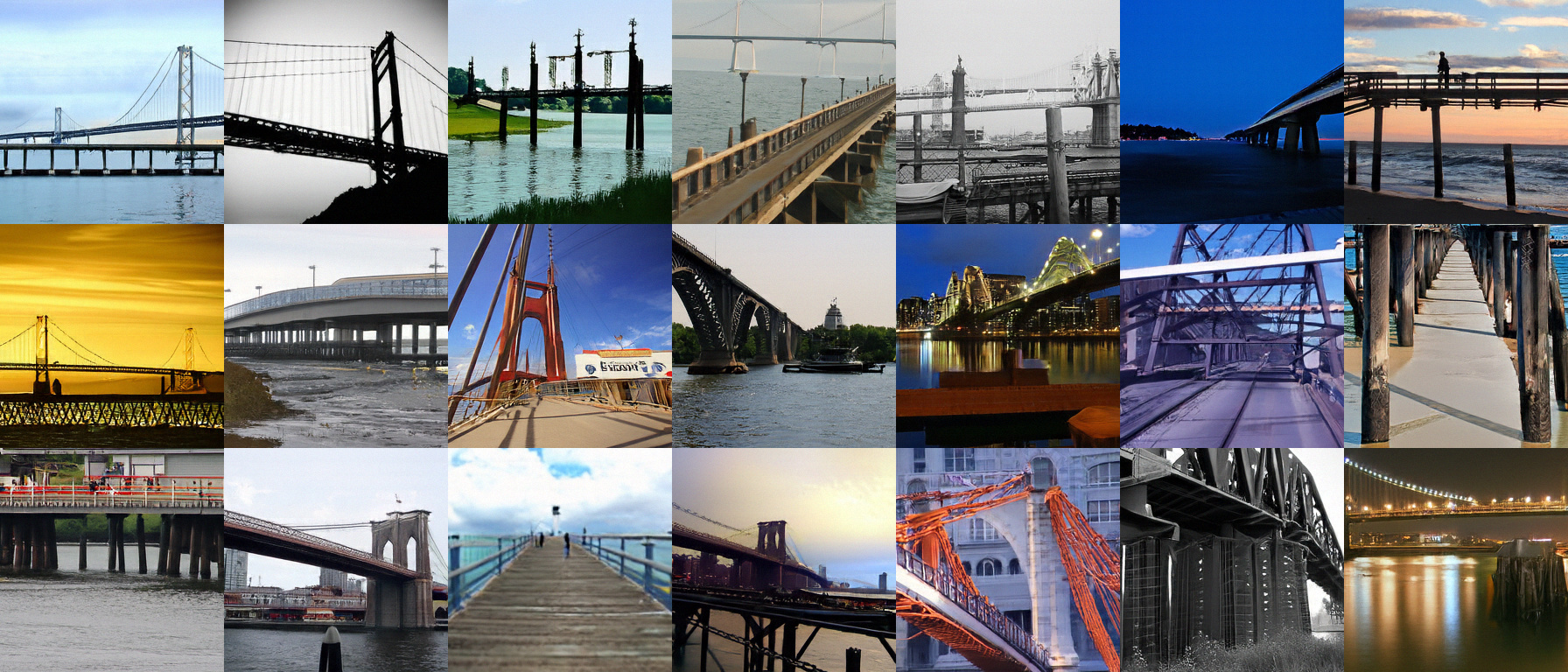}
  \par\vspace{-0.25em}
  {\scriptsize class 718: pier\par}
  \end{minipage}\hfill
  \begin{minipage}[t]{0.495\textwidth}
  \centering
  \includegraphics[width=\linewidth]{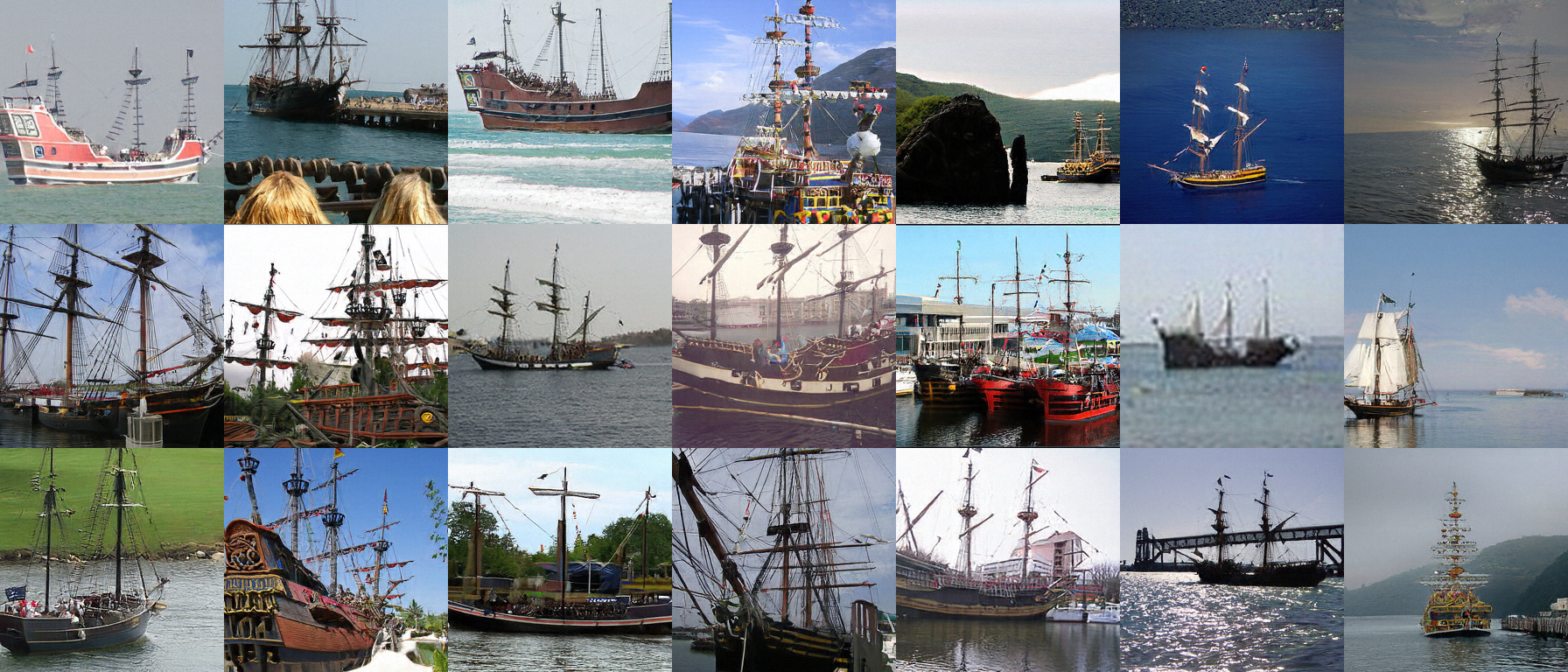}
  \par\vspace{-0.25em}
  {\scriptsize class 724: pirate, pirate ship\par}
  \end{minipage}

  \vspace{0.15em}

  \begin{minipage}[t]{0.495\textwidth}
  \centering
  \includegraphics[width=\linewidth]{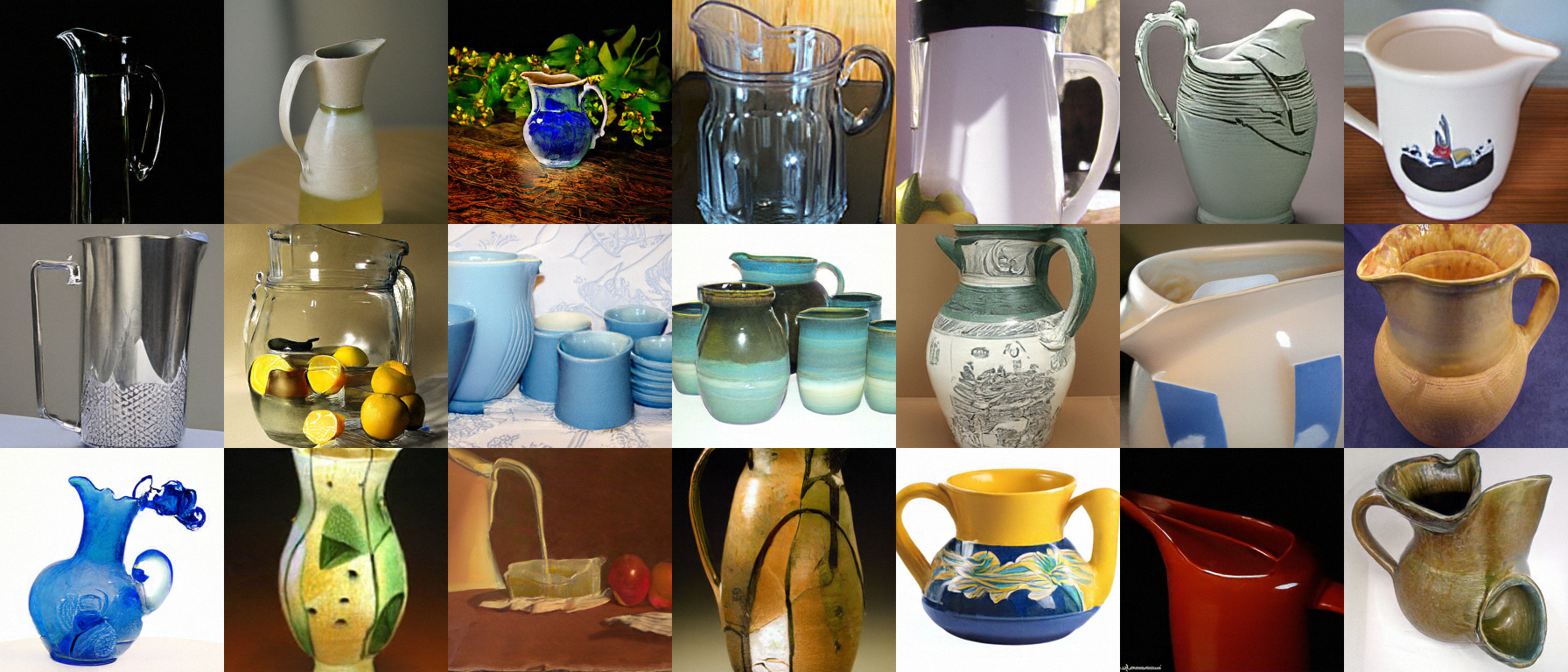}
  \par\vspace{-0.25em}
  {\scriptsize class 725: pitcher, ewer\par}
  \end{minipage}\hfill
  \begin{minipage}[t]{0.495\textwidth}
  \centering
  \includegraphics[width=\linewidth]{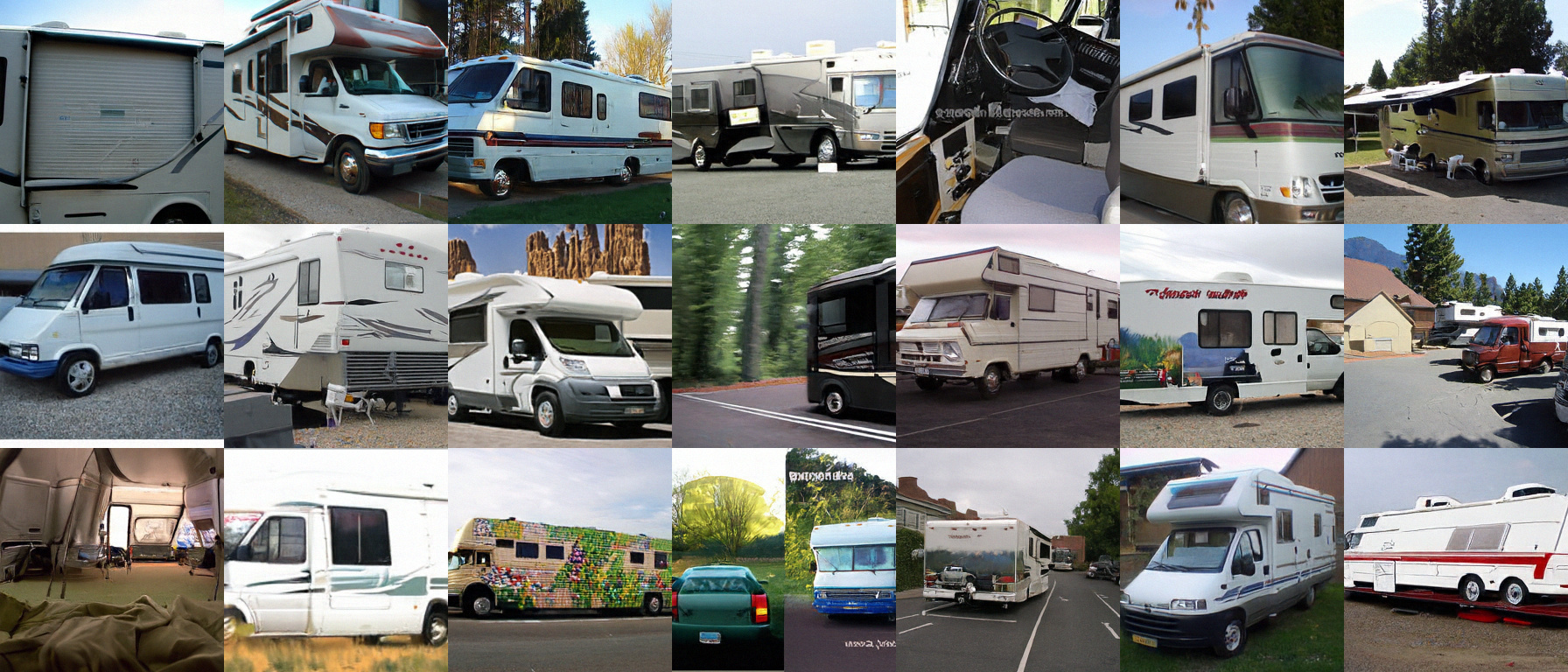}
  \par\vspace{-0.25em}
  {\scriptsize class 757: recreational vehicle, RV, R.V.\par}
  \end{minipage}

  \vspace{0.15em}

  \begin{minipage}[t]{0.495\textwidth}
  \centering
  \includegraphics[width=\linewidth]{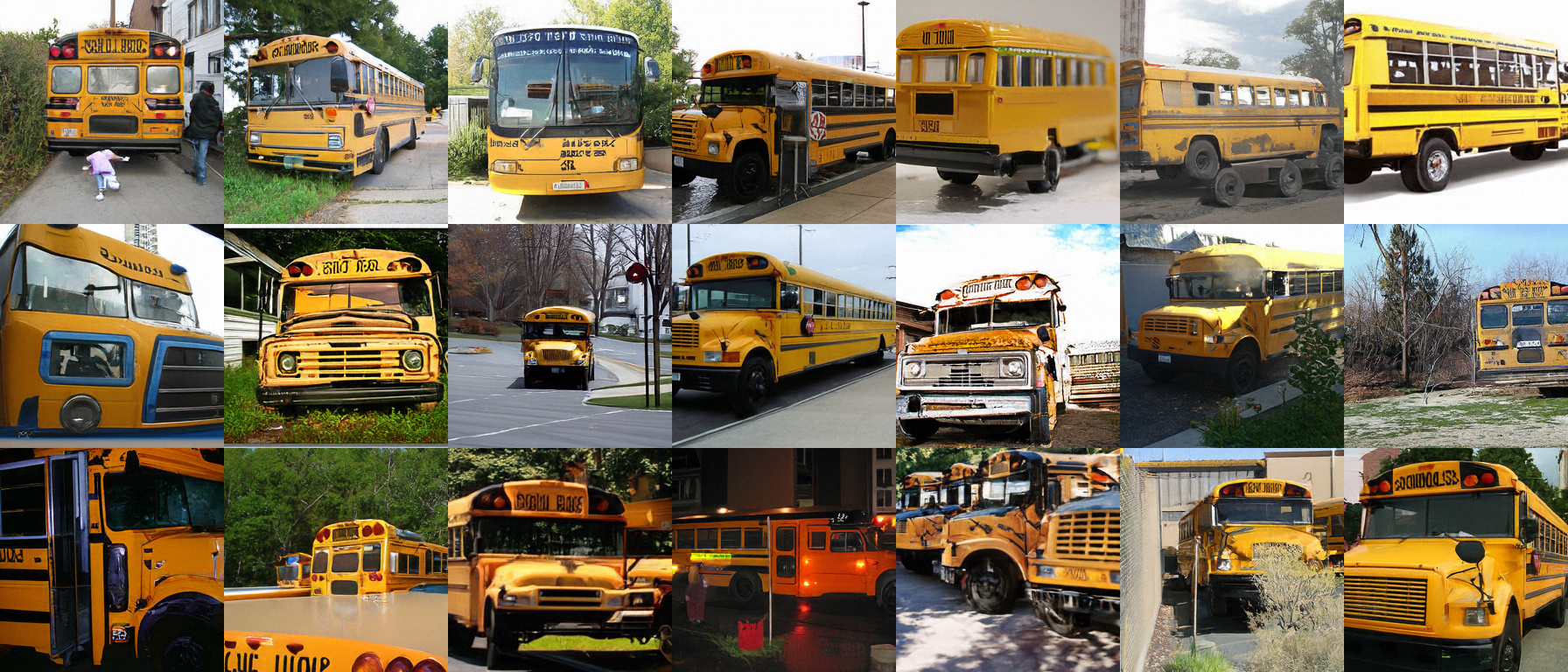}
  \par\vspace{-0.25em}
  {\scriptsize class 779: school bus\par}
  \end{minipage}\hfill
  \begin{minipage}[t]{0.495\textwidth}
  \centering
  \includegraphics[width=\linewidth]{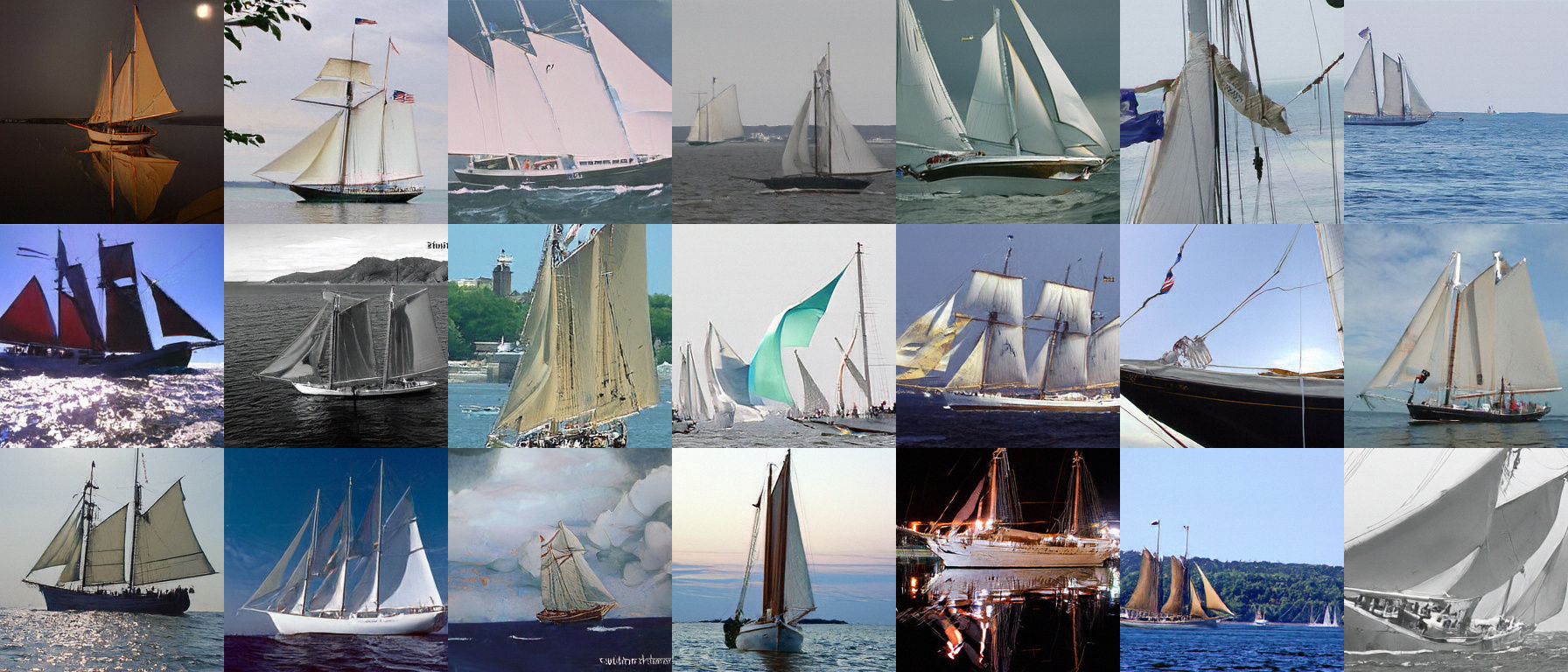}
  \par\vspace{-0.25em}
  {\scriptsize class 780: schooner\par}
  \end{minipage}

  \vspace{0.15em}

  \begin{minipage}[t]{0.495\textwidth}
  \centering
  \includegraphics[width=\linewidth]{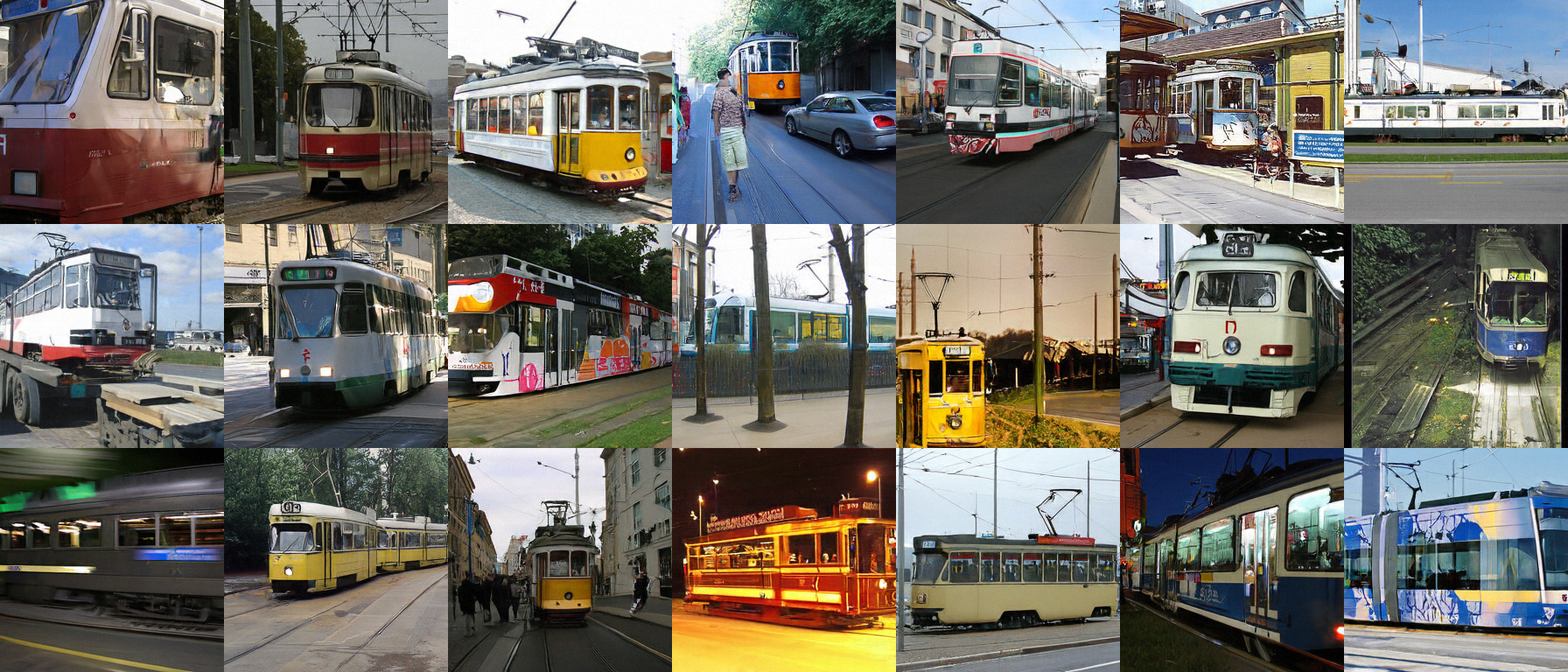}
  \par\vspace{-0.25em}
  {\scriptsize class 829: streetcar, tram, tramcar, trolley, trolley car\par}
  \end{minipage}\hfill
  \begin{minipage}[t]{0.495\textwidth}
  \centering
  \includegraphics[width=\linewidth]{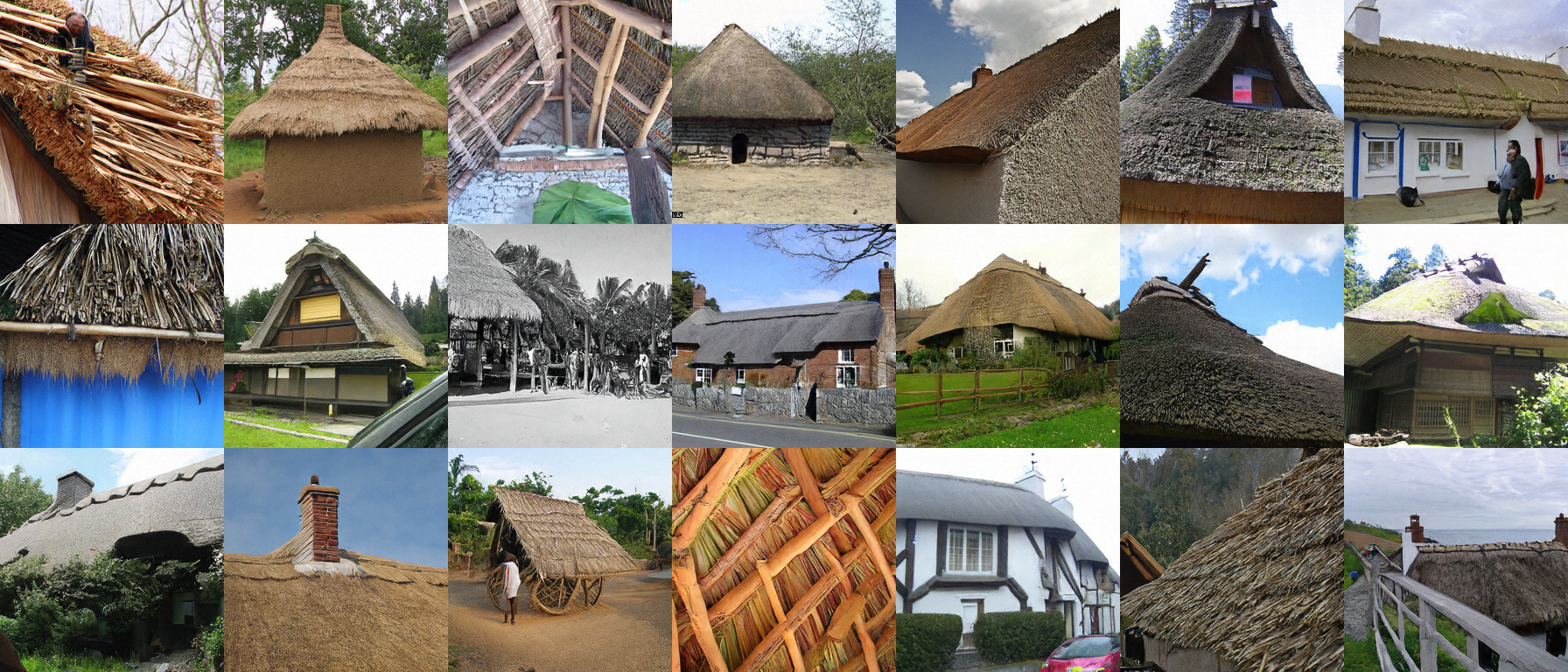}
  \par\vspace{-0.25em}
  {\scriptsize class 853: thatch, thatched roof\par}
  \end{minipage}

  \caption{Uncurated class-conditional samples on ImageNet $256\times256$ using PerF-H/16}
  \label{fig:perf_h16_samples_2}
\end{figure*}

\begin{figure*}[p]
  \centering

  \begin{minipage}[t]{0.495\textwidth}
  \centering
  \includegraphics[width=\linewidth]{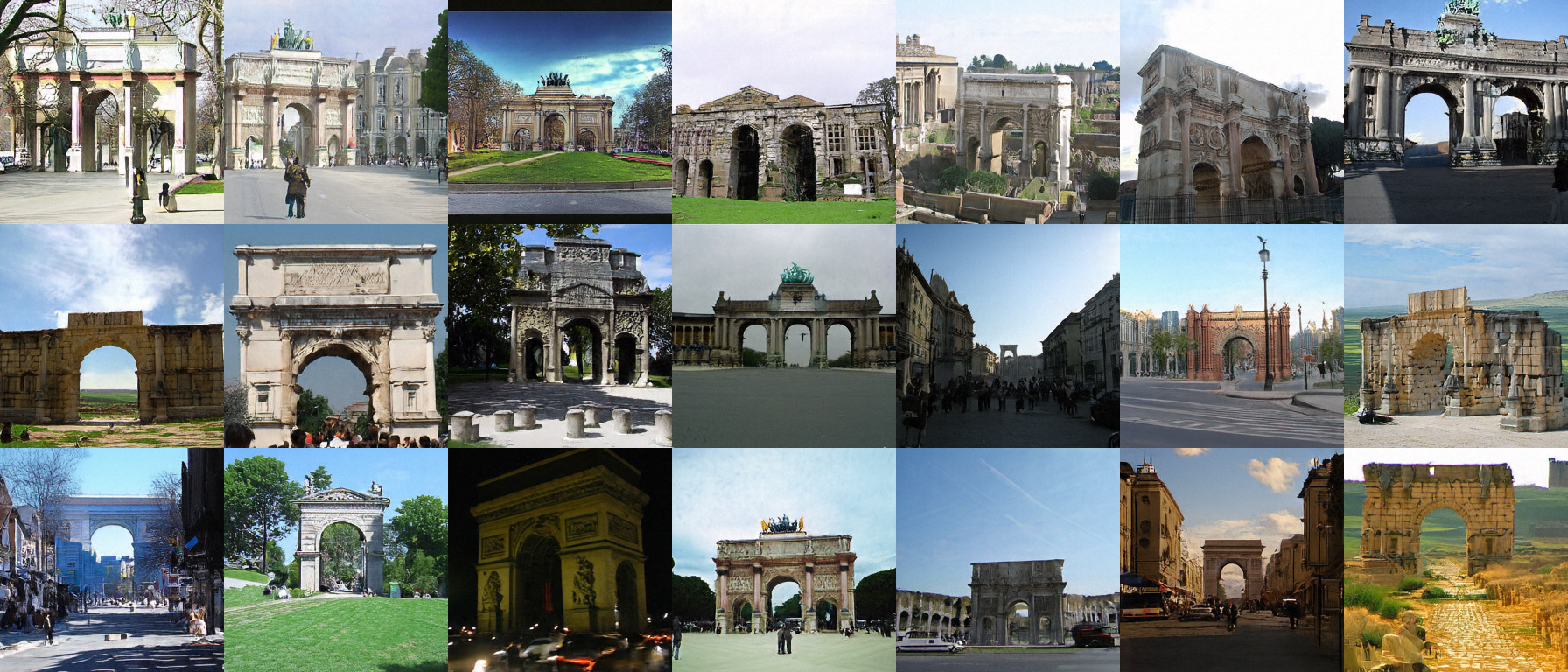}
  \par\vspace{-0.25em}
  {\scriptsize class 873: triumphal arch\par}
  \end{minipage}\hfill
  \begin{minipage}[t]{0.495\textwidth}
  \centering
  \includegraphics[width=\linewidth]{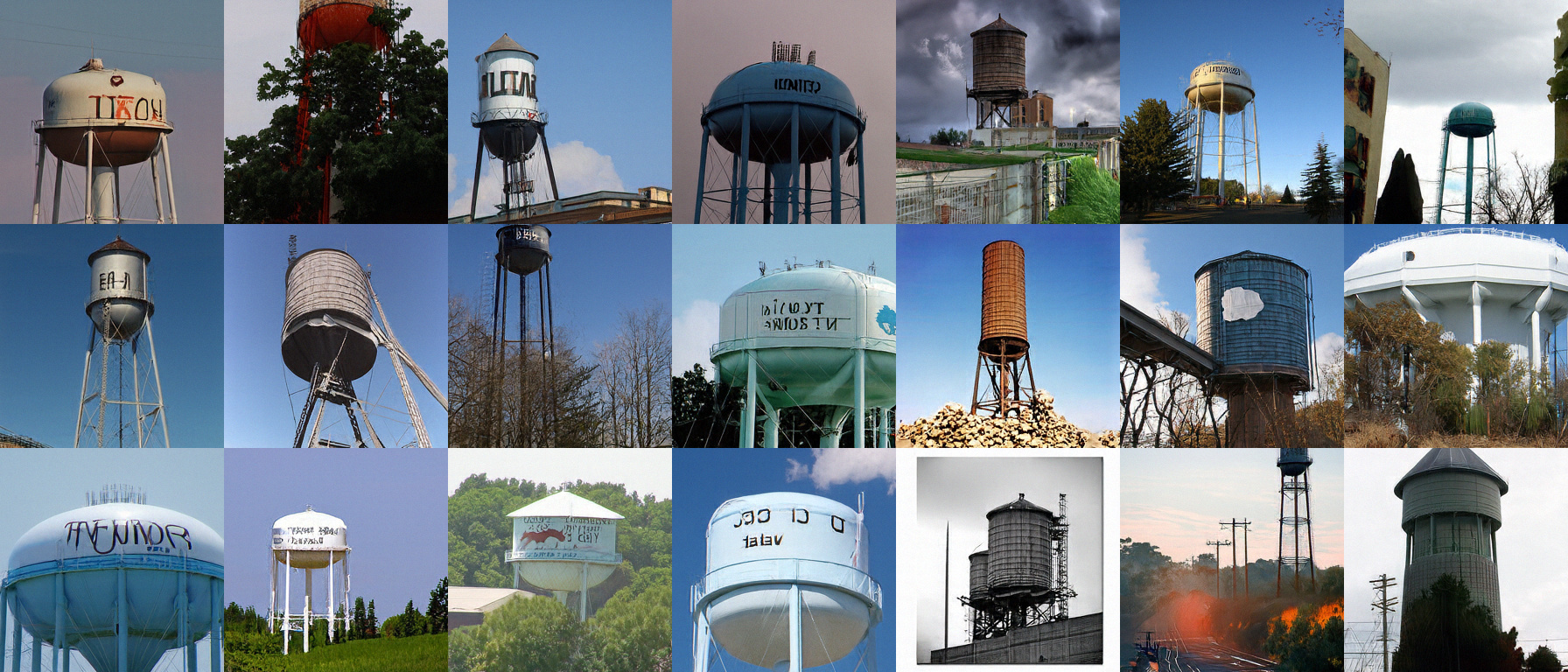}
  \par\vspace{-0.25em}
  {\scriptsize class 900: water tower\par}
  \end{minipage}

  \vspace{0.15em}

  \begin{minipage}[t]{0.495\textwidth}
  \centering
  \includegraphics[width=\linewidth]{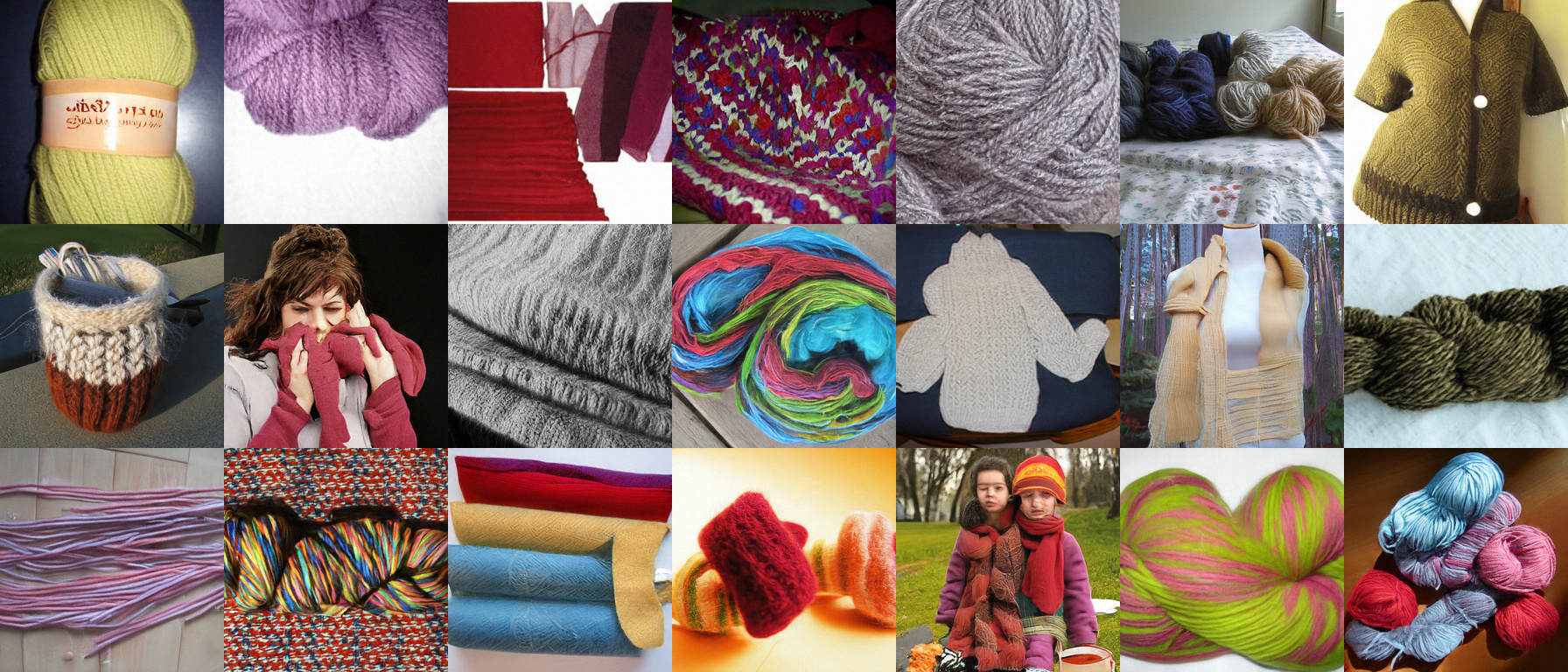}
  \par\vspace{-0.25em}
  {\scriptsize class 911: wool, woolen, woollen\par}
  \end{minipage}\hfill
  \begin{minipage}[t]{0.495\textwidth}
  \centering
  \includegraphics[width=\linewidth]{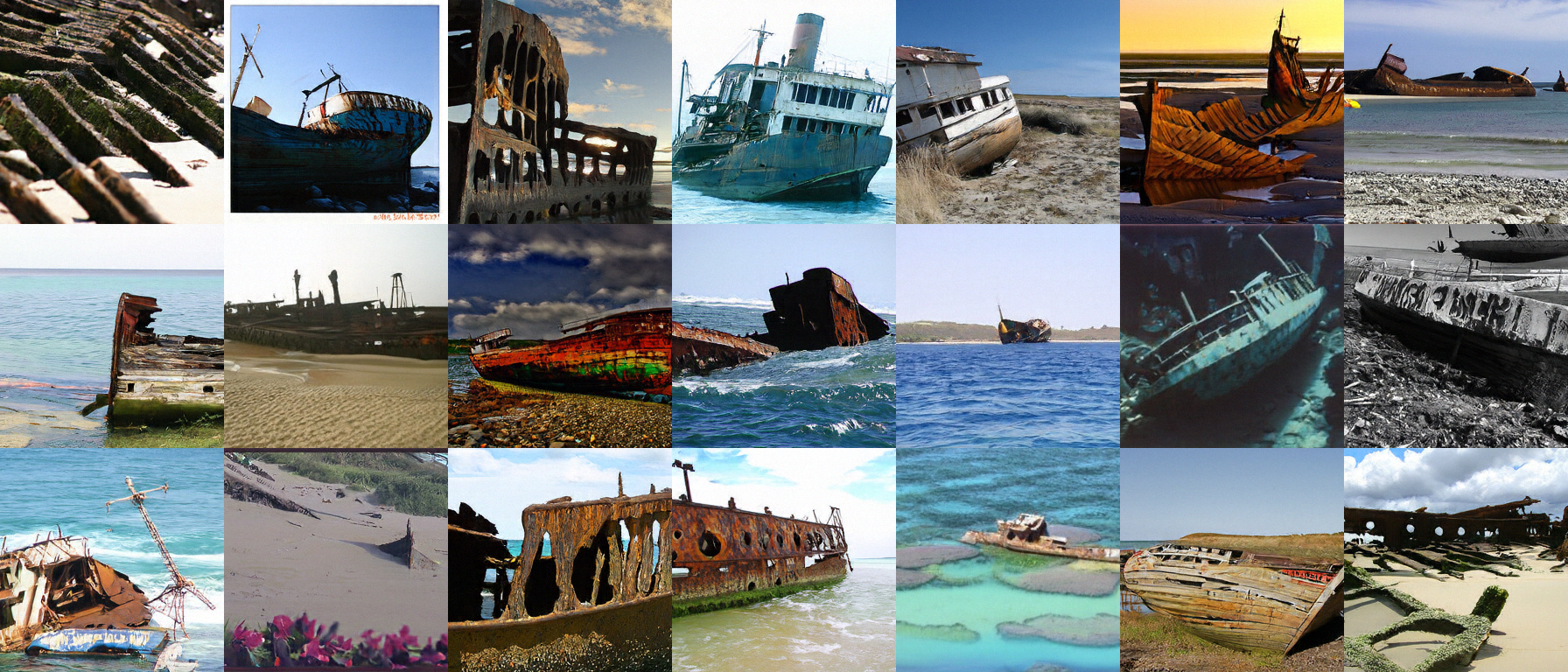}
  \par\vspace{-0.25em}
  {\scriptsize class 913: wreck\par}
  \end{minipage}

  \vspace{0.15em}

  \begin{minipage}[t]{0.495\textwidth}
  \centering
  \includegraphics[width=\linewidth]{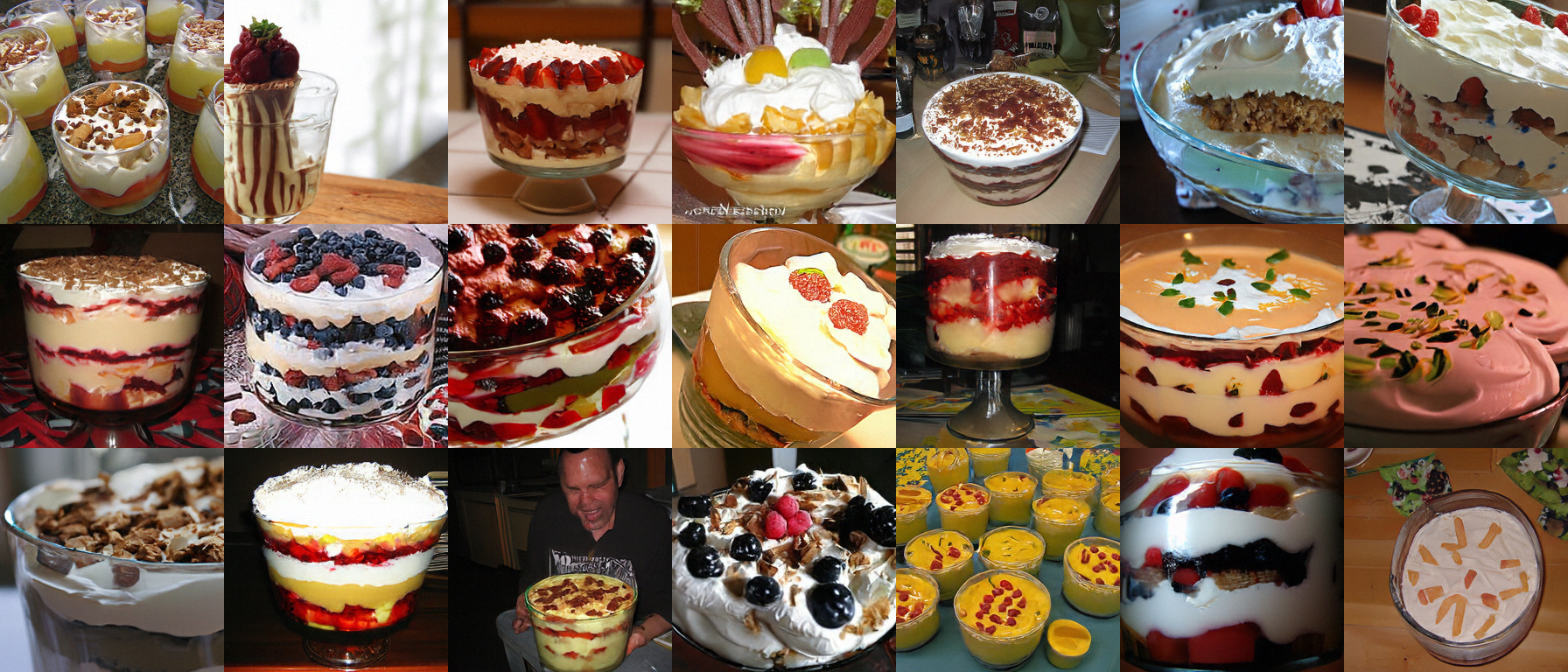}
  \par\vspace{-0.25em}
  {\scriptsize class 927: trifle\par}
  \end{minipage}\hfill
  \begin{minipage}[t]{0.495\textwidth}
  \centering
  \includegraphics[width=\linewidth]{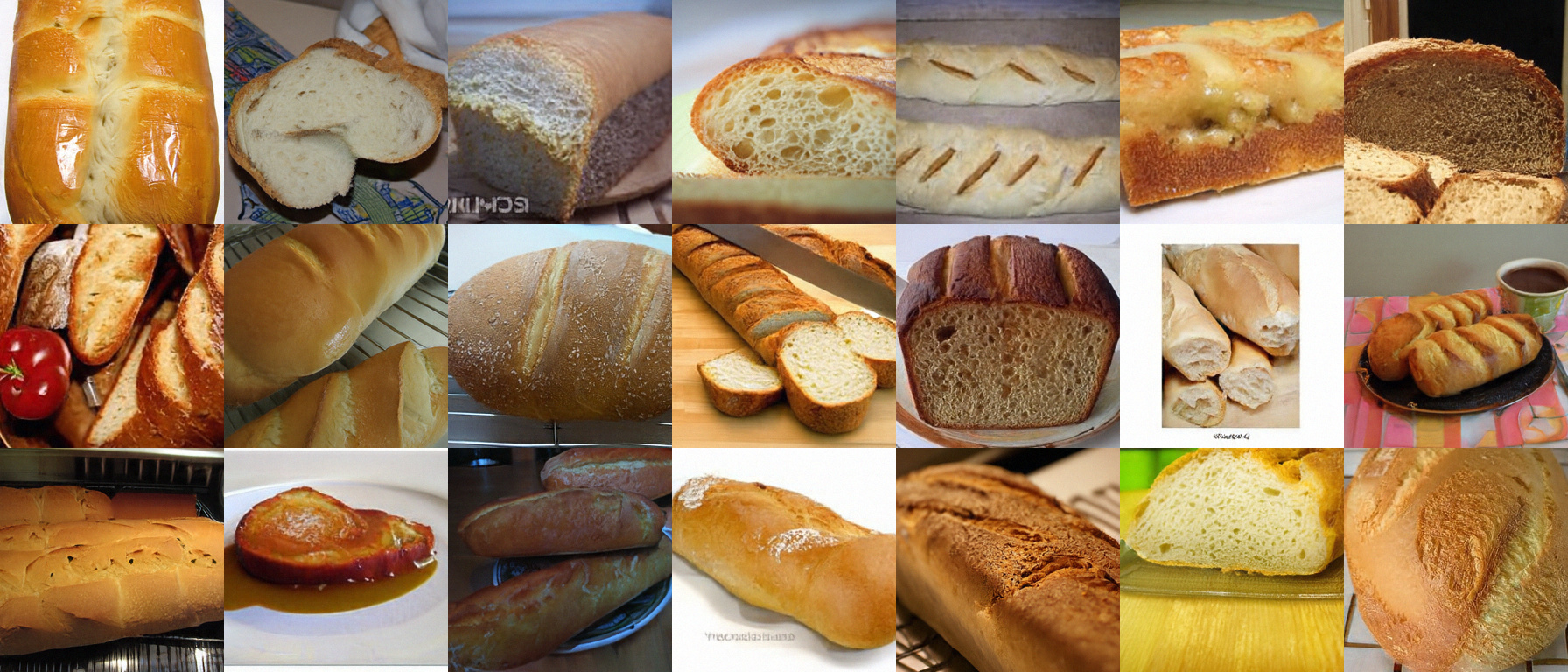}
  \par\vspace{-0.25em}
  {\scriptsize class 930: French loaf\par}
  \end{minipage}

  \vspace{0.15em}

  \begin{minipage}[t]{0.495\textwidth}
  \centering
  \includegraphics[width=\linewidth]{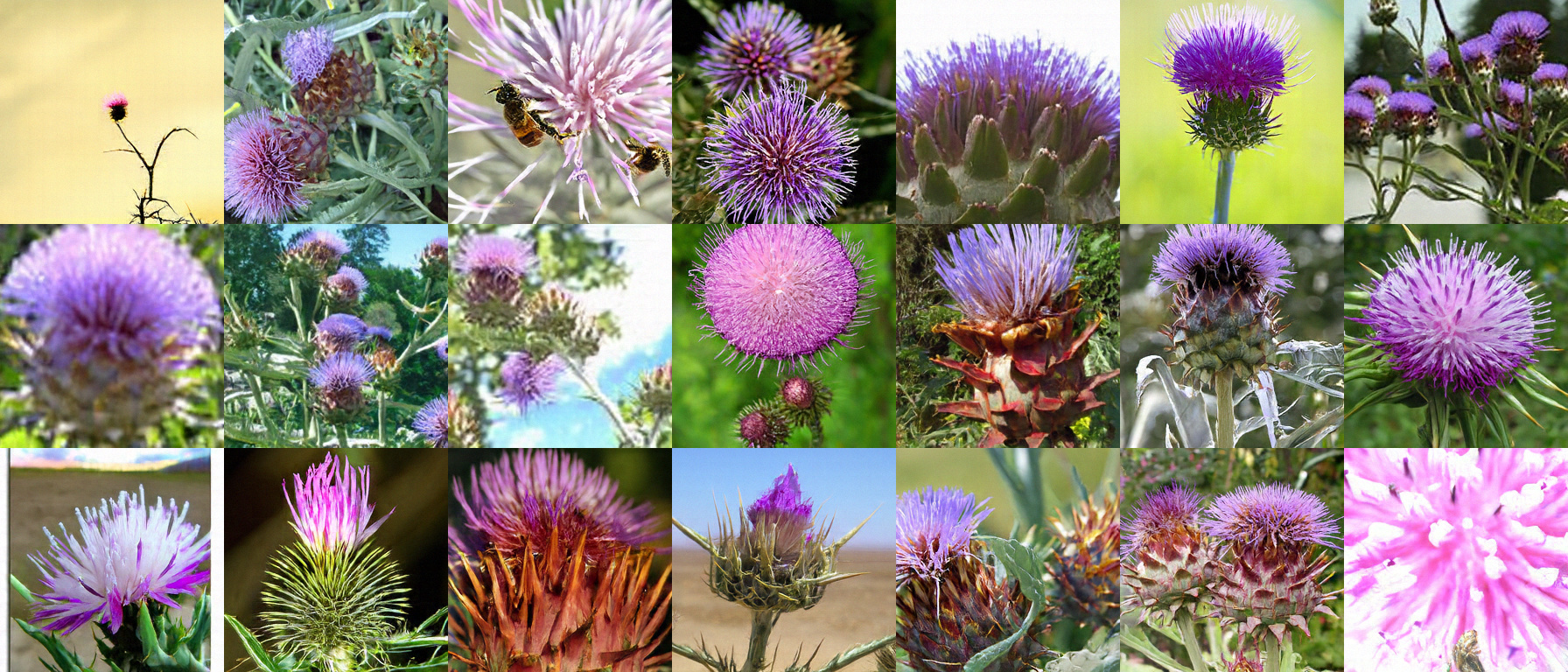}
  \par\vspace{-0.25em}
  {\scriptsize class 946: cardoon\par}
  \end{minipage}\hfill
  \begin{minipage}[t]{0.495\textwidth}
  \centering
  \includegraphics[width=\linewidth]{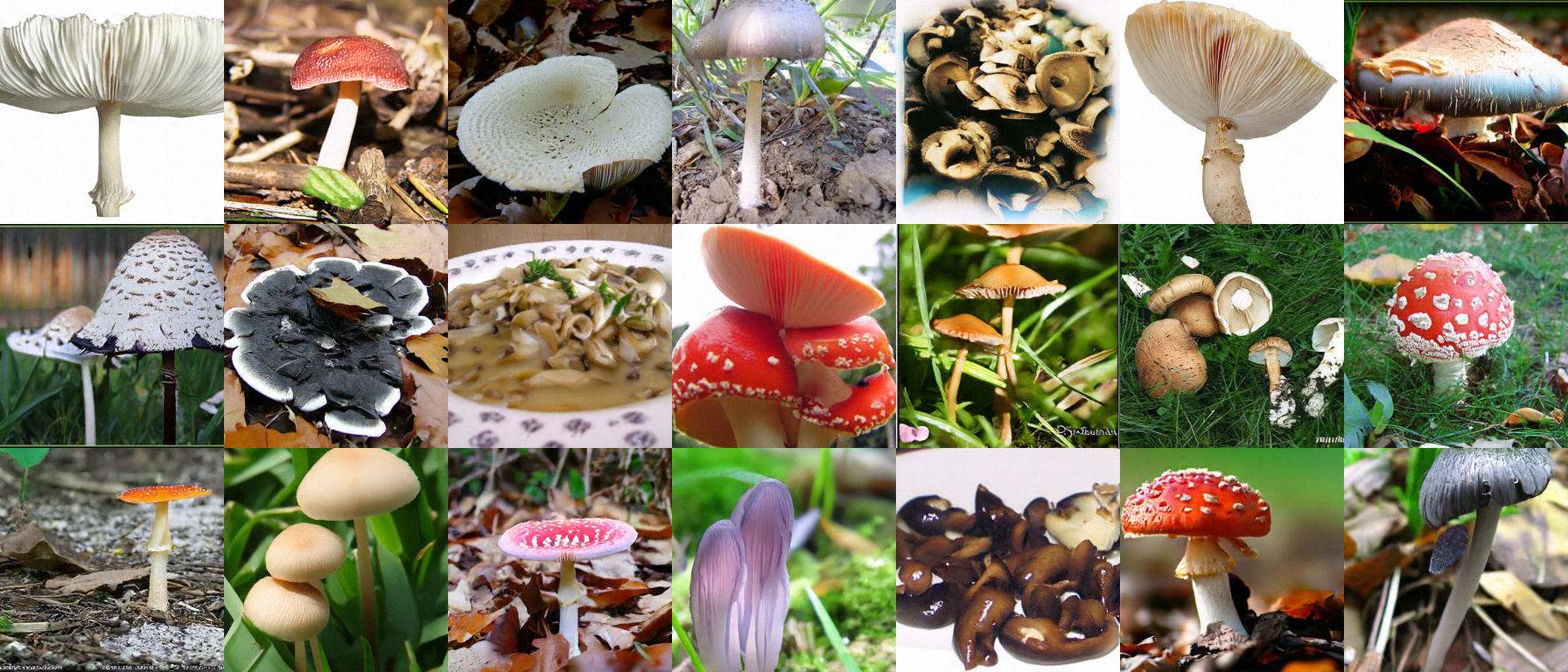}
  \par\vspace{-0.25em}
  {\scriptsize class 947: mushroom\par}
  \end{minipage}

  \vspace{0.15em}

  \begin{minipage}[t]{0.495\textwidth}
  \centering
  \includegraphics[width=\linewidth]{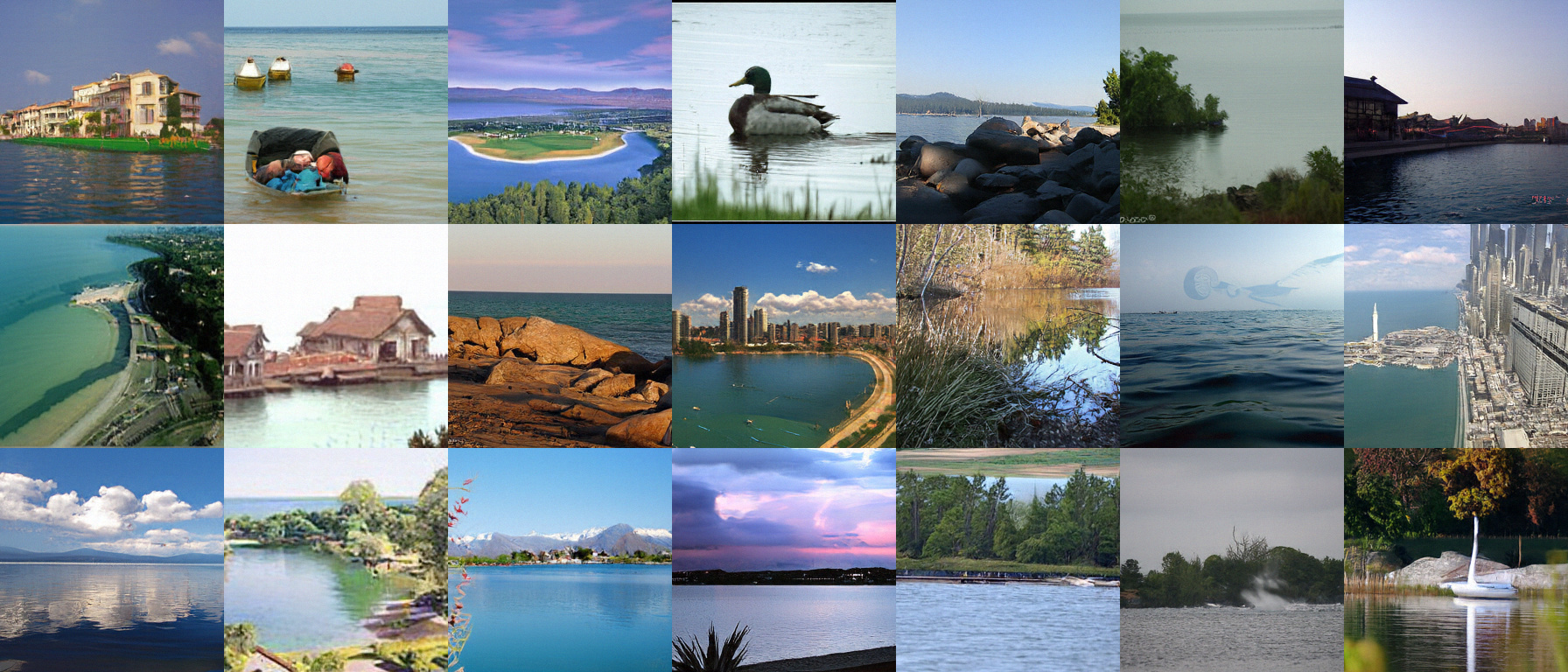}
  \par\vspace{-0.25em}
  {\scriptsize class 975: lakeside, lakeshore\par}
  \end{minipage}\hfill
  \begin{minipage}[t]{0.495\textwidth}
  \centering
  \includegraphics[width=\linewidth]{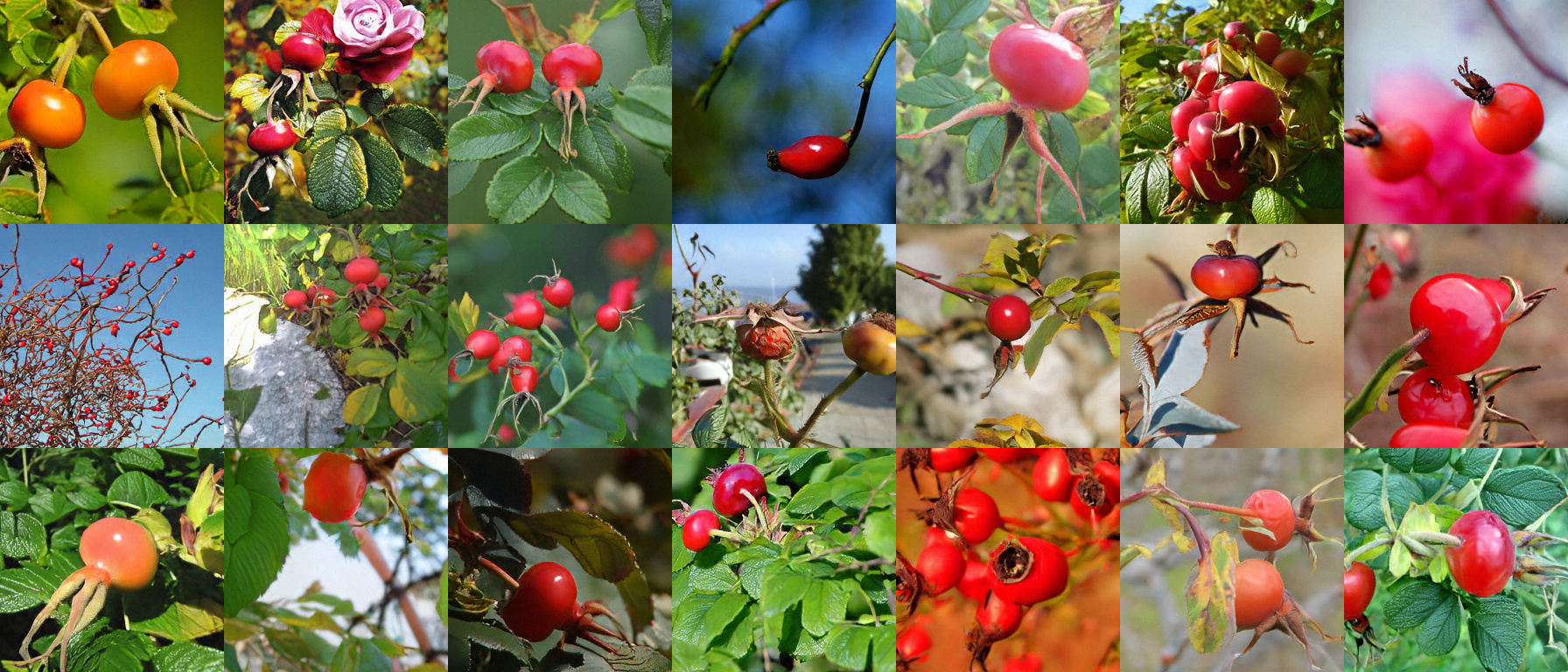}
  \par\vspace{-0.25em}
  {\scriptsize class 989: hip, rose hip, rosehip\par}
  \end{minipage}

  \caption{Uncurated class-conditional samples on ImageNet $256\times256$ using PerF-H/16.}
  \label{fig:perf_h16_samples_3}
\end{figure*}

We provide additional class-conditional samples generated by PerF-H/16. 
As seen in Figure~\ref{fig:perf_h16_samples_1}, Figure~\ref{fig:perf_h16_samples_2}, and Figure~\ref{fig:perf_h16_samples_3}, our method generates high-quality samples.

\end{document}

%% file: math_commands.tex
\usepackage{amsmath,amsfonts,bm}

\def\eqref#1{equation~\ref{#1}}

\def\1{\bm{1}}

\DeclareMathAlphabet{\mathsfit}{\encodingdefault}{\sfdefault}{m}{sl}
\SetMathAlphabet{\mathsfit}{bold}{\encodingdefault}{\sfdefault}{bx}{n}

